\documentclass[aos,MSNbibl,dvips]{arximspdf}
\usepackage{subeqn,subenv}
\usepackage{graphicx}
\usepackage{url,breakurl}

\doi{10.1214/13-AOS1187} 
\volume{42}
\issue{2}
\pubyear{2014}
\firstpage{532}
\lastpage{562}

\makeatletter
\newcommand{\rrVert}{\Vert}
\newcommand{\rrvert}{\vert}
\newcommand{\llVert}{\Vert}
\newcommand{\llvert}{\vert}
\def\cal{\mathcal}
\newcommand{\eqref}[1]{(\ref{#1})}
\newcommand{\T}{\mathcal{T}}
\newcommand{\V}{\mathcal{V}}
\newcommand{\N}{{\mathcal N}}
\newcommand{\Cond}{\kappa}
\newcommand{\const}{\mathsf{const}}
\newcommand{\off}{\mathrm{off}}
\newcommand{\ve}{\varepsilon}
\newcommand{\fnr}{\mathrm{FNR}}
\newcommand{\fpr}{\mathrm{FPR}}
\newcommand{\mcc}{\mathrm{MCC}}
\newcommand{\score}{\mathrm{score}}
\newcommand{\glasso}{\mathrm{glasso}}
\def\corr{\mathop{\operatorname{corr}}}
\newcommand{\event}{\mathcal{E}}
\newcommand{\diag}{\operatorname{diag}}

\def\hat{\widehat}
\def\tilde{\widetilde}
\newcommand{\tr}{\operatorname{tr}}

\def\argmin{\mathop{\operatorname{arg\,min}}}

\newcommand{\X}{{\mathcal X}}
\newcommand{\R}{\mathbf{R}}

\newcommand{\U}{\mathcal{U}}
\newcommand{\clime}{\mathrm{CL}}
\newcommand{\ind}{\mathbb{I}}

\newcommand{\degree}{\operatorname{deg}}
\newcommand{\init}{\operatorname{init}}
\newcommand{\e}{\epsilon}
\newcommand{\vp}{\varphi}
\newcommand{\A}{\mathcal{A}}

\newtheorem{theorem}{Theorem}[section]
\newproclaim{assumption}{Assumption}[section]

\newtheorem{lemma}[theorem]{Lemma}

\newproclaim{definition}[theorem]{Definition}
\newproclaim{example}[theorem]{Example}
\newproclaim{exm}[theorem]{Example}
\newproclaim{remark}[theorem]{Remark}
\newproclaim{rems}{Remarks}

\makeatother

\begin{document}
\begin{frontmatter}

\title{Gemini: Graph estimation with matrix variate normal instances}
\runtitle{Gemini}

\begin{aug}
\author{\fnms{Shuheng} \snm{Zhou}\corref{}\ead[label=e1]{shuhengz@umich.edu}\thanksref{t1}}
\thankstext{t1}{Supported in part by NSF Grant DMS-13-16731.}
\address{Department of Statistics\\
University of Michigan\\
Ann Arbor, Michigan 48109\\
USA\\
\printead{e1}}
\runauthor{S. Zhou}
\affiliation{University of Michigan}
\end{aug}

\received{\smonth{8} \syear{2013}}
\revised{\smonth{11} \syear{2013}}

%
\begin{abstract}
Undirected graphs can be used to describe matrix variate
distributions. In this paper, we develop new methods for
estimating the graphical structures and underlying parameters, namely,
the row and column covariance and inverse covariance matrices
from the matrix variate data.
Under sparsity conditions, we show that one is able to recover the
graphs and covariance matrices with a single random matrix
from the matrix variate normal distribution.
Our method extends, with suitable adaptation,
to the general setting where replicates are available.
We establish consistency and obtain the rates of convergence
in the operator and the Frobenius norm. We show that
having replicates will allow one to estimate more complicated
graphical structures and achieve faster rates of convergence.
We provide simulation evidence showing that
we can recover graphical structures as well as estimating
the precision matrices, as predicted by theory.
\end{abstract}


\end{frontmatter}

\section{Introduction}
\label{sec::intro}
The matrix variate normal model
has a long history in psychology and social sciences,
and is becoming increasingly popular in biology and genetics,
econometric theory, image and signal processing and machine learning in
recent years.
In this paper, we present a theoretical framework to show that
one can estimate the covariance and inverse covariance matrices well
using only one matrix from the matrix-variate normal distribution.
The motivation for this problem comes from many applications in
statistics and
machine learning. For example, in microarray studies, a single
$f \times m$ data matrix $X$ represents expression levels for $m$ genes
on $f$ microarrays; one needs to find out simultaneously the
correlations and partial
correlations between genes, as well as between microarrays.
Another example concerns observations from a spatiotemporal
stochastic process which can be described with a matrix normal distribution
with a separable covariance matrix $S \otimes T$, where typically, $S$
is called spatial covariance, $T$ is called the temporal
covariance and $\otimes$ is the Kronecker product.
When the stochastic process is spatial--temporal, some
structures can be assumed for one or both of the matrices in the
Kronecker product.
However, typically one has only one observational matrix.

We call the random matrix $X$ which contains $f$ rows and $m$
columns a single data matrix, or one instance from the matrix variate
normal distribution.
We say that an $f \times m$ random matrix $X$ follows a
matrix normal distribution with a separable covariance matrix $\Sigma
= A \otimes B$, which we write
%
\begin{equation}
\label{eq::matrix-normal-rep-intro} X_{f \times m} \sim\N_{f,m}(M, A_{m \times m} \otimes
B_{f \times f}).
\end{equation}
This is equivalent to say
$\operatorname{vec} \{ X  \}$ follows a multivariate normal
distribution with mean
$\operatorname{vec} \{ M  \}$ and covariance $\Sigma= A
\otimes B$. Here, $\operatorname{vec} \{ X  \}$ is formed
by stacking the columns of $X$ into a vector in $\R^{mf}$.
Intuitively, $A$ describes the covariance between columns of $X$
while $B$ describes the covariance between rows of $X$.
See~\cite{Dawid81,GV92} for characterization and examples.
Note that we can only estimate $A$
and $B$ up to a scaled factor, as $A \eta\otimes\frac{1}{\eta}B = A
\otimes B$ for any $\eta> 0$, and hence this will be our goal
of the paper, and precisely what we mean, when we say we are
interested in estimating covariances $A$ and $B$.

Undirected graphical models are often used to describe
high dimensional distributions.
We will use such descriptions in the present work
to encode structural assumptions on the inverse of the
row and column covariance matrices.
A common structural assumption is that
the inverse covariance matrices, also known as the precision matrices,
are sparse, which means that
the number of nonzero entries (sparsity levels) in one or both of them are
bounded. Under sparsity assumptions, a popular approach to obtain a sparse
estimate for the precision matrix is given by the $\ell_1$-norm
regularized maximum-likelihood
function, also known as the GLasso~\cite{YL07,FHT07,BGA08,RBLZ08}.
All these methods and their analysis assume that one is given
independent samples and the estimation of $A$ or
$B$ alone is their primary goal, as they all assume that $X$ has
either independent rows or independent columns.
A~direct application of the GLasso estimator to estimate $A \otimes B$ with
no regard for its separable structure will lead to computational
misery, as the cost will become prohibitive for $f, m$ in the order of
100. Various work~\cite{Dut99,LZ05,WJS08}
focused on algorithms and convergence properties on estimating
$\Sigma$ using a large number of samples $X(1), \ldots, X(n)$.
A mean-restricted matrix-variate normal model was considered in~\cite{AT10},
where they proposed placing additive penalties on estimated inverse
covariance matrices
in order to obtain regularized row and column covariance/precision matrices.
Other recent work with an iterative approach for solving the
graphical model selection problem in the context of matrix variate normal
distribution include~\cite{ZS10,YL12,LT12,THZ13,KLLZ13}.
None of these works was able to show convergence in the operator norm
which works in case $n=1$ and $f, m \to\infty$ as in our work.

\subsection{Our approach and contributions}
In this work, we take a penalized approach and
show from a theoretical point of view, the
advantages of estimating covariance matrices $A$, $B$ and the
graphs corresponding to their inverses simultaneously albeit via
separable optimization functions.
The key observation and starting point of our work is: although $A$ and
$B$ are not identifiable
given the separable representation as in~\eqref{eq::matrix-normal-rep-intro},
their correlation matrices $\rho(A)$ and $\rho(B)$, and
the graphical structures corresponding to their inverses are
identifiable, and can indeed be efficiently estimated for a given
matrix $X \sim\mathcal{N}_{f,m}(0, A \otimes B)$. Moreover,
$\rho(A)^{-1}$ and $\rho(B)^{-1}$ encode the same structural
information as $A^{-1}$ and $B^{-1}$ do, in the sense that they
share an identical set of nonzero edges.
Therefore, we propose estimating the overall ${\Sigma} = A \otimes B$
and its inverse by (i) first estimating correlation matrices $\rho(A)$
and $\rho(B)$
(and their inverses) using a pair of
$\ell_1$-norm penalized estimators for an instance $X \sim\mathcal
{N}_{f,m}(0,
A \otimes B)$,
(ii) and then combining these two estimators with the estimated
variances to form an estimator for ${\Sigma}$.

Toward this end, we develop Gemini (\underline{G}raph \underline{e}stimation
with \underline{m}atrix var\underline{i}ate \underline{n}ormal
\underline{i}nstances), a new method for
estimating graphical structures, and the underlying parameters $A$ and
$B$. We will answer the following question: how sparse
does $A^{-1}$ or $B^{-1}$ need to be in order for us to obtain
statistical convergence rates for estimating $A$ and $B$ (up to a
scaled factor)
simultaneously with one data matrix~${X}$?
Our estimators extend, with suitable adaptation, to the
general setting where $n$ replicates of $X$ are available.
Our method is computational efficient. The dominating cost involves
in estimating $\rho(A)^{-1}$ and $\rho(B)^{-1}$: the total cost is in
the order of $O(f^3 + m^3)$ for sparse graphs or $O(f^4 + m^4)$ for
general graphs.

In summary, we make the following theoretical contributions: (i)
consistency and
rates of convergence in the operator and the Frobenius norm of the covariance
matrices and their inverses, (ii) large deviation results for the sample
correlation estimators which we propose for estimating both the row and column
correlation and covariance matrices given a single matrix or multiple
replicates of the matrix-normal
data, (iii) conditions that guarantee simultaneous estimation of the
graphs for
both rows and columns.
We note that with all other parameters hold invariant, the rates of convergence
in all metrics in (i) and (ii) in estimating $A$, $B$ (and their
inverses) will be proportional to $n^{-1/2}$.
To the best of our knowledge, these are the first such
results on the matrix-variate normal distributions in the high dimensional
setting for finite and small sample instances, by which we mean $n
<\log\max(m,f)$.
We provide simulation evidence and a real data example showing that we
can recover graphical structures as well as estimate the precision
matrices effectively.

There is no known closed-form solution for the maximum of the
likelihood function for the matrix-variate normal distribution.
There has been a line of work in the literature which suggested using
iterative algorithms, namely, the Flip-Flop methods to
estimate the covariance matrix with the Kronecker structure; see, for
example,~\cite{Dut99,LZ05,WJS08} and references therein.
In the present work, building upon the baseline Gemini estimators,
we also propose a three-step penalized variant of the Flip-Flop algorithms
in Section~\ref{sec::MLE}. We show that under an additional condition,
this approach yields certain improvements upon the baseline Gemini
estimators.

The rest of the paper is organized as follows.
In Section~\ref{sec::methods}, we will define our model and the method.
Section~\ref{sec::main} presents the main theoretical results in this
paper on estimating $A \otimes B$, as well as discussions on our
method and results; moreover, we review the related work to place our
work in context.
Section~\ref{sec::gemini-est} provides large deviation inequalities
for the sample correlation coefficients in approximating the
underlying parameters of $\rho(A)$ and $\rho(B)$;
more general bounds of this nature are derived in
Section~13 in the supplementary material~\cite
{Zhou13supp}.
Convergence rates in the Frobenius norm for estimating the inverse
correlation matrices are also derived.
We propose a Noniterative Penalized Flip-Flop algorithm and
study its convergence properties in Sections~\ref{sec::MLE} and~\ref
{sec::flip-flop}.
Section~\ref{sec::numerical} shows our numerical results.
We conclude in Section~\ref{sec::conclude}.
We place all technical proofs in the supplementary material~\cite{Zhou13supp}.

\subsection{Notation}
For a matrix $A = (a_{ij})_{1\le i,j\le m}$, let $\llVert  A\rrVert _{\max} =
\max_{i,j} |a_{ij}|$ denote
the entry-wise max norm; let $\llVert  A\rrVert _{1} = \max_{j}\sum_{i=1}^m\llvert  a_{ij}\rrvert $
denote the matrix $\ell_1$ norm.
The Frobenius norm is given by $\llVert  A\rrVert ^2_F = \sum_i\sum_j a_{ij}^2$.
Let $|A|$ denote the determinant and $\operatorname{ tr}(A)$ be the trace of $A$.
Let $\varphi_{\max}(A)$ and $\varphi_{\min}(A)$ be the
largest and smallest eigenvalues, and $\kappa(A)$ be the condition
number for matrix $A$.
The operator or $\ell_2$ norm $\llVert  A\rrVert _2^2$ is
given by
$\varphi_{\max}(AA^T)$.
Let $r(A)= {\llVert  A\rrVert _F^2 }/{\llVert  A\rrVert
_2^2}$ denote the stable rank for
matrix $A$.
We write $|\cdot|_1$ for the $\ell_1$ norm of a matrix vectorized,
that is, $|A|_1 = \llVert \operatorname{vec} \{ A  \}
\rrVert _1 = \sum_{i}\sum_j |a_{ij}|$.
Let $\llvert A\rrvert _{1,\off} = \sum_{i \neq j} |A_{ij}|$, and
$\llvert A\rrvert _{0,\off}$ be the number of nonzero nondiagonal
entries in the
matrix. We use $A^{-T}$ to denote $(A^{-1})^T$.
We write $\diag(A)$ for a diagonal matrix with the same diagonal as~$A$.
For a symmetric matrix $A$, let $\Upsilon(A) =  (\upsilon
_{ij} )$
where $\upsilon_{ij} = \ind(a_{ij} \neq0)$, where $\mathbb
{I}(\cdot)$ is the indicator function.
Let $I$ be the identity matrix.
We let $C$ be a constant which may change from line to line.
For two numbers $a, b$, $a \wedge b := \min(a, b)$, and
$a \vee b := \max(a, b)$.
We write $a \asymp b$ if $ca \le b \le Ca$ for some positive absolute
constants $c,C$ which are independent of $n, f, m$ or sparsity
parameters.

\section{The model and the method}
\label{sec::methods}
In the matrix variate normal setting, we aim to
estimate the row and column covariance (correlation) matrices, from which
we can obtain an estimate for $\Sigma$.
The problem of covariance estimation in the context of matrix variate
normal distribution
is intimately connected to the problem of graphical model selection,
where the graphs corresponding to the column and the row vectors
are determined by the sparsity patterns (or the zeros) of $B^{-1}$ and
$A^{-1}$, respectively.
Graph estimation in this work means precisely the estimation of the
zeros, as well as the
nonzero entries in $A^{-1}$ and $B^{-1}$.
We formulate such correspondence precisely in
Section~\ref{sec::matrix-graphs}.
We define our estimators in Sections~\ref{sec::gemini-estimators}
and \ref{sec::gemini-rep}.

\subsection{Problem definition: The matrix normal graphical model}
\label{sec::matrix-graphs}
\begin{figure}

\includegraphics{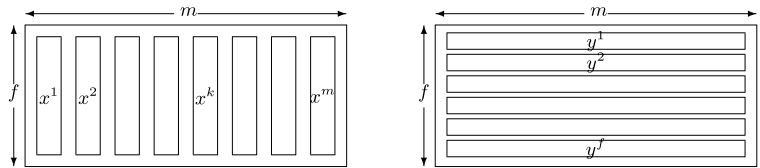}

\caption{Column and row vectors of matrix $X$, where $X \sim
\mathcal{N}_{f,m}(0, A \otimes B)$. Let $A = (a_{ij})$ and $B = (b_{ij})$.
The normalized column vectors ${x^{1}}/{\sqrt{a_{11}}}, \ldots,
x^{m}/\sqrt{a_{mm}}$, where $a_{ii} > 0$, follow a multivariate normal
distribution $\mathcal{N}_f(0, B)$ while normalized row vectors
$y^{1}/\sqrt{b_{11}},\ldots, y^{f}/\sqrt{b_{ff}}$, where $b_{jj} >
0$, follow $\mathcal{N}_m(0, A)$.}
\label{fig::columns}
\end{figure}

We show in Figure~\ref{fig::columns} the data matrix $X$ and its
column vectors: $x^1, x^2, \ldots, x^{k}, \ldots, x^{m}$,
and row vectors $y^1, y^2, \ldots, y^f$.
%

%
%
This notation is followed throughout the rest of the paper.
First recall the following definition concerning the
classical Gaussian graphical model for a random vector.
%
\begin{definition}
\label{def::ggm}
Let $V = (V_1, \ldots,V_f)^T$ be a random Gaussian vector,
which we represent by an undirected graph $G = (\mathcal{V}, F)$. The
vertex set
$\mathcal{V} := \{1, \ldots, f\}$ has one vertex for each component
of the vector $V$. The edge set
$F$ consists of pairs $(j, k)$ that are joined by an edge. If $V_j$ is
independent of $V_k$ given the other variables, then $(j, k) \notin F$.
\end{definition}
Now let $\V=\{1, \ldots, f\}$ be an index set which enumerates
rows of $X$ according to a fixed order.
For all $i=1, \ldots, m$, we assign to each variable of a column vector
$x^i$ exactly one element of the set $\V$ by a rule of
correspondence $g\dvtx x^i \to\V$ such that $g(x^i_j) = j, j=1,\ldots, f$.
The graphs $G_i(\V, F)$ constructed
for each random column vector $x^i, i=1, \ldots, m$ according to
Definition~\ref{def::ggm}
will share an identical edge set $F$,
because the normalized column vectors ${x^{1}}/{\sqrt{a_{11}}},
\ldots,
x^{m}/\sqrt{a_{mm}}$ follow the same multivariate normal distribution
$\N_f(0, B)$.
Hence, graphs $G_1, \ldots, G_m$ are isomorphic and we write $G_i
\simeq G_j, \forall i, j$.
Due to the isomorphism,
we use $G(\V, F)$ to represent the family of graphs $G_1, \ldots, G_m$.
Hence, a pair $(\ell, k)$ which is absent in $F$ encodes
conditional independence between the ${\ell}${th} row and the
$k$th row give all other rows.
Similarly, let $\Gamma=\{1, \ldots, m\}$ be the index set
which enumerates columns of $X$ according to a fixed order.
We use $H(\Gamma, E)$ to represent the family of graphs $H_1, \ldots,
H_f$, where $H_i$ is constructed for row vector $y^i$, and
$H_i \simeq H_j, \forall i, j$.
Now $H(\Gamma, E)$ is a graph with adjacency matrix $\Upsilon(H) =
\Upsilon(A^{-1})$ as edges in $E$ encode nonzeros in $A^{-1}$.
And $G(\V, F)$ is a graph with adjacency matrix
$\Upsilon(G) = \Upsilon(B^{-1})$.
The Kronecker product, $H \otimes G$, is defined as the graph with adjacency
matrix
$\Upsilon(H) \otimes\Upsilon(G)$~\cite{Weich62},
where clearly missing edges correspond to zeros in the inverse
covariance $A^{-1} \otimes B^{-1}$, and $H \otimes G$ represents the
graph of the $p$-variate Gaussian random vector $\operatorname
{vec} \{ X  \}$, where $p = mf$.
In the present work, we aim to estimate $\Upsilon(H)$ and $\Upsilon
(G)$ separately.
Estimating their Kronecker product directly following the classical
$p$-variate Gaussian graphical
modeling approach will be costly in terms of both computation and the
sample requirements.

\subsection{The Gemini estimators}
\label{sec::gemini-estimators}
We start with the one-matrix case.
We note that between $A$ and $B$, the dimension of one matrix
is the same as the number of samples available for estimating
parameters in the
other matrix in case $n=1$. Therefore, $m$ and $f$ are allowed to grow
so long as
they grow with respect to each other.
The first hurdle we need to deal with, besides the simultaneous row
and column correlations, is the fact that between the two covariance
matrices $A$ and $B$
(as well as their inverses),
the one with the higher dimension, which contains more canonical
parameters, is always left with a smaller number of correlated samples
in order to
achieve its inference tasks.
The remedy comes from the following observation.
Although ambient dimension $f, m$ cannot be both bounded
by the other unless $f=m$, the sparsity over nondiagonal
entries of each precision matrix can be assumed to be
bounded by the ambient dimension of the other.

Under such sparsity assumptions, we first provide a pair of separable
regularized estimators for the correlation matrices
$\rho(A) =  ({a_{ij}}/{\sqrt{a_{ii} a_{jj}}} )$ and
$\rho(B) =  ({b_{ij}}/{\sqrt{b_{ii} b_{jj}}} )$,
\begin{subeqnarray}
\hat{A}_{\rho}
& = & \argmin_{A_{\rho} \succ0} \bigl \{\operatorname{ tr}\bigl(\hat{\Gamma}(A)
A_{\rho}^{-1}\bigr) +
\log\llvert  A_{\rho}\rrvert  + \lambda_{B}\bigl \llvert A_{\rho
}^{-1}\bigr\rrvert _{1,\off}  \bigr\},\label{eq::A-Penest-corr-intro}
\\
\label{eq::B-Penest-corr-intro}
\hat{B}_{\rho}
& = & \argmin_{B_{\rho} \succ0}  \bigl\{\operatorname{ tr}\bigl(\hat{\Gamma}(B)
B_{\rho}^{-1}\bigr) +
\log\llvert  B_{\rho}\rrvert  + \lambda_{A} \bigl\llvert B_{\rho
}^{-1}\bigr\rrvert _{1,\off}  \bigr\},
\end{subeqnarray}
where the input are a pair of sample correlation matrices
$\hat{\Gamma}(A)$ and $\hat{\Gamma}(B)$
%
\begin{equation}
\label{eq::sample-corr-A-intro} \hat\Gamma_{ij}(A) := \frac{\langle{ x^i, x^j }
\rangle}{\llVert  x^i\rrVert _2{\llVert  x^j\rrVert _2}} \quad\mbox{and}\quad
\hat\Gamma_{ij}(B) := \frac{\langle{ y^i, y^j }\rangle}{\llVert  y^i\rrVert
_2{\llVert  y^j\rrVert _2}},
\end{equation}
and the $\ell_1$ penalties are imposed on the off-diagonal
entries of the inverse correlation estimates.
Note that the population parameters $A$ and $B$ can be written as
\[
A \otimes B:= \bigl(W_1 \rho(A) W_1 \bigr) \otimes
\bigl(W_2 \rho(B) W_2 \bigr)/\bigl(\tr(A)\tr(B)\bigr),
\]
where ${W}_1/\sqrt{\tr(B)} = \diag(\sqrt{a_{11}}, \ldots,
\sqrt{a_{mm}})$ and ${W}_2/\sqrt{\tr(A)} = \diag(\sqrt{b_{11}},\break
\ldots,
\sqrt{b_{ff}})$.
In order to get an estimate for $A \otimes B$,
we multiply each of the two regularized estimators $\hat{A}_{\rho}$
and $\hat{B}_{\rho}$
by an estimated weight matrix $\hat{W}_1$ or $\hat{W}_2$, respectively,
\begin{eqnarray*}
\label{eq::hatW1-orig} \hat{W}_1 & = & \diag\bigl(\bigl\llVert
x^1\bigr\rrVert _2, \bigl\llVert x^2\bigr
\rrVert _2, \ldots, \bigl\llVert x^m\bigr\rrVert
_2\bigr) = \diag\bigl(X^T X\bigr)^{1/2},
\\
\label{eq::hatW2-orig} \hat{W}_2 & = & \diag\bigl(\bigl\|y^1
\bigr\|_2,\bigr \|y^2\bigr\|_2, \ldots,\bigl \|y^f
\bigr\|_2\bigr) = \diag\bigl(X X^T \bigr)^{1/2}.
\end{eqnarray*}
Up to a multiplicative factor $\tr(B)$ and $\tr(A)$,
$\hat{W}_1^2$ and $\hat{W}_2^2$ will provide an estimate for
$\diag(A)$ and $\diag(B)$, respectively;
hence, to estimate $A \otimes B$, we compute the Kronecker product of
our weighted
estimators,
\[
\hat{A \otimes B}  :=  (\hat{W}_1 \hat{A}_{\rho} \hat
{W}_1 ) \otimes (\hat{W}_2 \hat{B}_{\rho}
\hat{W}_2 )/\llVert X\rrVert _F^2
\]
while adjusting the unknown multiplicative factors
$\tr(B)\tr(A)$ by $\llVert  X\rrVert _F^2$.

Clearly, the sample correlation estimators
\eqref{eq::sample-corr-A-intro} are obtained from the gram matrices
$X^T X$ and $X X^T$ of the column and row vectors as follows:
%
\begin{eqnarray}\qquad\quad
 \label{eq::matrix-normal-est-A} \hat\Gamma(A) &=& \hat{W}_1^{-1}
\bigl(X^TX\bigr) \hat{W}_1^{-1} \quad\mbox{and}\quad
\hat\Gamma(B) = \hat{W}_2^{-1}\bigl(XX^T\bigr)
\hat{W}_2^{-1} \qquad\mbox{where}
\\
\mathbb{E}X^T X &=& \sum_{i=1}^f
\mathbb{E} {y^i \otimes y^i} = \tr(B)A,\qquad \mathbb{E} {X
X^T} = \sum_{i=1}^m \mathbb{E}
{x^i \otimes x^i} = \tr(A)B,
\end{eqnarray}
and the multiplicative factors $\tr(B)$ and $\tr(A)$
become irrelevant due to cancellation.
By setting the gradient equations of objective
functions~\eqref{eq::A-Penest-corr-intro} and
\eqref{eq::B-Penest-corr-intro} to zero, we see that the pair of
estimators satisfy $\diag(\hat{A}_{\rho}) = \diag(\hat{\Gamma
}(A))$ and
$\diag(\hat{B}_{\rho}) = \diag(\hat{\Gamma}(B))$ as desired.
Moreover, the penalty parameters $\lambda_{B}$ and $\lambda_{A}$
are chosen to
dominate the maximum of
entry-wise errors for estimating $\rho(A)$ and $\rho(B)$
with $\hat{\Gamma}(A)$ and $\hat{\Gamma}(B)$ as characterized in
Theorem~\ref{thm::large-devi-cor}
(cf. Remark~\ref{rem::CA} and the comments which follow immediately).

\subsection{Gemini for replicates of $X$}
\label{sec::gemini-rep}
We now adapt the Gemini
estimators as defined in Section~\ref{sec::gemini-estimators} to the
general setting where we have multiple replicates of~$X$.
Suppose that we have $n$ independently and identically distributed matrices
$X(1), \ldots, X(n) \sim\mathcal{N}_{f,m}(0, A \otimes B)$.
For each $t$, we denote by
%
\begin{equation}
\label{eq::vectors} X(t) = \bigl[\matrix{x(t)^1 &x(t)^2 &\cdots&
x(t)^{m}}\bigr] = \bigl[\matrix{y(t)^1 &y(t)^2 &\cdots&
y(t)^f}\bigr]^T
\end{equation}
the matrix $X_{f \times m}(t)$ with $x(t)^1, \ldots, x(t)^m \in
\R^{f}$ being its columns vectors and $y^1(t), \ldots, y^f(t)$ being
its row vectors.

First, we update our sample correlation matrices,
which we will plug
in~\eqref{eq::A-Penest-corr-intro} and~\eqref{eq::B-Penest-corr-intro}
to obtain the penalized correlation estimators $\hat{A}_{\rho}$ and
$\hat{B}_{\rho}$.
\begin{subeqnarray}
\hat\Gamma_{ij}(A) & := &
\frac{\sum_{t=1}^n \langle{ x(t)^i, x(t)^j }\rangle}{
\sqrt{\sum_{t=1}^n \llVert  x(t)^i\rrVert _2^2}
\sqrt{\sum_{t=1}^n\llVert  x(t)^j\rrVert _2^2}},\label{eq::sample-cor-Arep} \\
\label{eq::sample-cor-Brep}
\hat\Gamma_{ij}(B) & := &
\frac{\sum_{t=1}^n \langle{ y(t)^i, y(t)^j }\rangle} {
\sqrt{\sum_{t=1}^n \llVert  y(t)^i\rrVert _2^2}
\sqrt{\sum_{t=1}^n {\llVert  y(t)^j\rrVert _2}^2}}.
\end{subeqnarray}
Next, we update the weight matrices $\hat{W}_1$ and $\hat{W}_2$ as follows:
\begin{subeqnarray}
\hat{W}_1
& = &\diag \Biggl(
\sqrt{\frac{1}{n}\sum_{t=1}^n \bigl\llVert  x(t)^{i}\bigr\rrVert
_2^2}, i=1, \ldots,
m \Biggr),\label{eq::hatW1} \\
\label{eq::hatW2}
\hat{W}_2
& = &
\diag \Biggl(\sqrt{\frac{1}{n}\sum_{t=1}^n \bigl\llVert  y(t)^{j}\bigr\rrVert _2^2},
j=1,\ldots, f \Biggr).
\end{subeqnarray}
We can then construct an estimator for $A \otimes B$ as before,
%
\begin{equation}
\label{eq::ABestimate} \hat{A \otimes B} :=  (\hat{W}_1
\hat{A}_{\rho}\hat{W}_1 ) \otimes (\hat{W}_2
\hat{B}_{\rho} \hat{W}_2 )\Big/ \Biggl(\frac{1}{n}\sum
_{t=1}^n \bigl\llVert X(t)\bigr\rrVert
_F^2 \Biggr).
\end{equation}
We will show in Theorem~\ref{thm::large-devi-cor-multiple}
large deviation bounds for estimating the correlation coefficients in
$\rho(A)$ and $\rho(B)$
with entries in sample correlation $\hat\Gamma(A)$ and $\hat\Gamma(B)$
constructed above, which are crucial in proving
the convergence rates for estimating $A \otimes B$ and its inverse
with $\hat{A \otimes B}$ and $\hat{A \otimes B}^{-1}$.

\section{Theoretical results}
\label{sec::main}
In this section, we present in Theorem~\ref{thm::main} and
Theorem~\ref{thm::main-II} the convergence rates
for estimating the row and column covariance matrices and their
inverses with respect to the operator norm and the Frobenius norm,
respectively.
Our analysis is nonasymptotic in nature; however, we first formulate our
results from an asymptotic point of view for simplicity.
To do so, we consider an array of matrix variate normal data
%
\begin{equation}
\label{data} X(1),\ldots,X(n) \mbox{ i.i.d.} \sim{\cal N}_{f,m}(0,A_0
\otimes B_0),\qquad n = 1,2,\ldots,
\end{equation}
where $f, m$ may change with $n$.
Let $\llvert A_0^{-1}\rrvert _{0,\off}$
and $\llvert B_0^{-1}\rrvert _{0,\off}$ be the number of nonzero nondiagonal
entries in the inverse covariance matrices $A_0^{-1}$ and $B_0^{-1}$,
respectively. Recall, for matrix $A$, $r(A)
={\llVert  A\rrVert _F^2}/{\llVert  A\rrVert _2^2}$
and $\kappa(A)$ denote its stable rank and condition number, respectively.

We make the following assumptions.
\begin{longlist}[(A1)]
\item[(A1)]
The dimensions $f$ and $m$ are allowed to grow with respect to each
other, and
\begin{eqnarray*}
\label{eq::sparA} \bigl\llvert A_0^{-1}\bigr\rrvert
_{0,\off} & = & o \bigl(nf/ \log(m \vee f) \bigr)\qquad (f,m \to \infty)\quad\mbox{and}
\\
\label{eq::sparB}  \bigl\llvert B_0^{-1}\bigr
\rrvert _{0,\off} & = & o \bigl(nm/ \log (m \vee f) \bigr)\qquad (f,m \to\infty).
\end{eqnarray*}
\item[(A2)]
The eigenvalues $\varphi_i(A_0), \varphi_j(B_0), \forall i, j$ of
the positive definite covariance matrices $A_0$ and
$B_0$ are bounded away from $0$ and $+\infty$.

Moreover, we assume that the stable ranks $r(A_0)$ and $r(B_0)$
satisfy
$r(A_0),\break   r(B_0) \geq4 \log(m \vee f)/n$, which
holds trivially if $n \ge4 \log(m \vee f)$;
otherwise, it is sufficient to require that
$(m \vee f) =o (\exp (\frac{f}{\kappa^2(B_0)} \wedge
\frac{m}{\kappa^2(A_0)} ) )$.
\end{longlist}
We now state the main results of this paper, which are new to the best
of our knowledge.
These bounds are stated in terms of the relative errors.
%
\begin{theorem}
\label{thm::main}
Consider data generating random matrices in (\ref{data}).
Suppose that \textup{(A1)} and \textup{(A2)} hold, and
the penalty parameters are chosen to be
\begin{eqnarray*}
\lambda_{A} = \lambda_{A_0} & \asymp& \frac{\llVert  A_0\rrVert _F}{\tr(A_0)}
\frac{\log^{1/2} (m
\vee f)}{\sqrt{n}} \asymp\frac{\log^{1/2} (m \vee f)}{\sqrt{mn}} \to0\quad\mbox{and}
\\
 \lambda_{B} = \lambda_{B_0} & \asymp&
\frac{\llVert  B_0\rrVert _F}{\tr(B_0)} \frac{\log^{1/2} (m \vee f)}{\sqrt{n}} \asymp\frac{\log^{1/2} (m \vee f)}{\sqrt{fn}} \to0.
\end{eqnarray*}
Then with probability at least $1-\frac{3}{(m \vee f)^2}$,
for $\hat{A \otimes B}$ as defined in~\eqref{eq::ABestimate},
%
\begin{eqnarray}
 \llVert \hat{A \otimes B} - A_0 \otimes B_0\rrVert
_2 &\leq&\llVert A_0\rrVert _2\llVert
B_0\rrVert _2 \delta\quad \mbox{and}
\nonumber\\
 \bigl\llVert \hat{A\otimes B}^{-1} - A_0^{-1}
\otimes B_0^{-1}\bigr\rrVert _2 &\leq &\bigl
\llVert A_0^{-1}\bigr\rrVert _2 \bigl\llVert
B_0^{-1}\bigr\rrVert _2 \delta'\nonumber\\
\eqntext{\mbox{where } \displaystyle\delta, \delta' = O \bigl(\lambda_{A_0}\sqrt
{\bigl\llvert B_0^{-1}\bigr\rrvert _{0,\off}
\vee1} + \lambda_{B_0} \sqrt{\bigl\llvert A_0^{-1}
\bigr\rrvert _{0,\off} \vee1} \bigr) = o(1).}
\end{eqnarray}
\end{theorem}

\begin{theorem}
\label{thm::main-II}
Consider data generating random matrices as in (\ref{data}).
Let $\lambda_{A_0}$ and $\lambda_{B_0}$
be chosen as in Theorem~\ref{thm::main}.
Let $\hat{A \otimes B}$ be as defined in~\eqref{eq::ABestimate}.
Under~\textup{(A1)} and \textup{(A2)},
%
\begin{eqnarray}\label{eq::frob-rate}
\llVert \hat{A \otimes B} - A_0 \otimes
B_0\rrVert _F \leq\delta\llVert A_0\rrVert
_F \llVert B_0\rrVert _F
\nonumber
\\[-8pt]
\\[-8pt]
\eqntext{\mbox{where } \delta= O \bigl(\lambda_{A_0} \sqrt{\bigl
\llvert B_0^{-1}\bigr\rrvert _{0,\off} \vee f}/
\sqrt{f} + \lambda_{B_0} \sqrt{\bigl\llvert A_0^{-1}
\bigr\rrvert _{0,\off} \vee m }/\sqrt{m} \bigr) = o(1).}
\end{eqnarray}
In particular, suppose
\textup{(i)} $1 \leq n \leq\log(m \vee f)$ or \textup{(ii)}
$\llvert A_0^{-1}\rrvert _{0,\off} =O( m)$ and $\llvert B_0^{-1}\rrvert _{0,\off} = O(f)$.
Then
$\delta=O( \lambda_{A_0} + \lambda_{B_0})$.
The same conclusions hold for the inverse estimate,
with $\delta$ being bounded in the same order as in \eqref{eq::frob-rate}.
\end{theorem}
The two summands in $\delta$ and $\delta'$ in Theorems~\ref{thm::main}
and~\ref{thm::main-II} correspond to the rates of convergence in
the operator and the Frobenius norm for estimating the row and column
covariance matrices $A_0, B_0$, up to a scale factor, respectively.
These rates are derived in Sections~\ref{sec::inverse-frob}
and~10.
We prove Theorems~\ref{thm::main} and~\ref{thm::main-II}
in Section~11 in the supplementary material~\cite
{Zhou13supp},
where we examine the rate of \eqref{eq::frob-rate} in case $n \ge
4\log(m \vee f)$ in
Remark~11.3, and show the absolute error bounds in
Theorems~11.1 and~11.2.
There we also make the connection between the one-matrix and the
multiple-matrix cases in order to understand the rates for $n > 1$.

\subsection{Discussion}
\label{sec::discussion}
To put our discussions on the rates of convergence for covariance
estimation in context, we first present an example from the classical
multivariate analysis.
Consider the case where we are given a single
sample from the matrix variate normal\vadjust{\goodbreak} distribution with $B_0 = I$,
and the dimensions $f, m$ increase to infinity, while
the aspect ratio $f/m \to\const>1$.
The classical multivariate analysis focuses on estimating $A_0$
using data matrix $X$;
the simplest way to estimate $A_0$ is to compute the sample covariance
\[
\tilde{A}_f = \frac{1}{f}X^T X =
\frac{1}{f}\sum_{i=1}^f
x_i \otimes x_i\qquad \mbox {where } x_1, \ldots,
x_m\ \mbox{i.i.d.} \sim\N_m(0, A_0).
\]
The problem here is to determine the
minimal number of independent rows we need so that the
sample covariance matrix $\tilde{A}_f$ approximates $A$ ``well'' in the
operator norm.
This concerns the classical ``Bai--Yin law'' in random matrix theory
regarding the Wishart random matrix $\tilde{A}_f$,
which says that the spectrum of $\tilde{A}_f$ is almost surely contained
in the interval $[a^2/f + o(1), b^2/f + o(1)]$ where $a = (\sqrt{f}
-\sqrt{m})_{+}$
and $b = \sqrt{f} +\sqrt{m}$ in case $A_0 = I$.
For general covariance matrix $A_0$, the following holds with high
probability (cf.~\cite{vers12}):
%
\begin{equation}
\label{eq:BaiYin} \Vert\tilde{A}_f - A_0
\Vert_2 \leq \bigl(2 \sqrt{m/f} + (m/f) + o(1) \bigr) \llVert
A_0\rrVert _2.
\end{equation}
While such results provide a satisfactory answer to the covariance estimation
problem in the regime $f \geq m$ for general multivariate normal
distributions, it remains challenging to answer the following questions:
(a) how to estimate the covariance matrix which has the larger
dimension of the two? That is, how can we approximate $A_0$ well in the
operator norm when $f < m$? (b) how to estimate both $A_0$ and $B_0$
given both correlated rows and columns?

Our answer to the first question is to use the penalized methods.
The operator norm bound in Theorem~\ref{thm::main} illustrates the
point that the combination of sparsity and spectral assumptions as in
(A1), (A2) and $\ell_1$-regularization ensures convergence on
estimation of the covariance and
precision matrices, even though their ambient dimensions
may greatly exceed the given sample sizes.
In particular, the ambient dimensions which appear in the numerator
in~\eqref{eq:BaiYin} are
replaced with the sparsity parameters (cf. Theorem~\ref{thm::main}):
\[
\label{eq::opnorm} \delta, \delta' = O \bigl(\log^{1/2} (m
\vee f) \bigl(\sqrt{\bigl\llvert A_0^{-1}\bigr
\rrvert _{0,\off} \vee1}/{\sqrt{f}} + \sqrt{\bigl\llvert
B_0^{-1}\bigr\rrvert _{0,\off} \vee1}/{\sqrt{m}}
\bigr) \bigr) = o(1),
\]
which holds for $n=1$ with high probability under (A1) and (A2), as
(A1) implies that, up to a logarithmic factor, the number of nonzero
off-diagonal entries in $A_0^{-1}$ or $B_0^{-1}$ must be bounded by the
dimension
of the other matrix. We will relax such sparsity conditions in
Section~\ref{sec::Uclass}.

To answer the second question,
first recall that in the current setting, \eqref
{eq::matrix-normal-est-A} suggests that $\tilde{A}_f = X^T X/f$ and
$\tilde{B}_m = X X^T/m$
are good starting points for us to construct estimators for $A_0, B_0$,
$\rho(A_0)$, and $\rho(B_0)$
despite the presence of dependence along the other dimension.
The relationships between the row and column correlations of $X$ are known
to complicate the solution to the related problem of testing
the hypothesis that microarrays are independent of each other
given
possibly correlated genes~\cite{Efr09}.\vspace*{1pt}
Taking these complex relationships into consideration,
we construct covariance and correlation
estimators based on $\tilde{A}_f$ and $\tilde{B}_m$, as well as
the pair of functions in \eqref{eq::alternate-A};\vadjust{\goodbreak} we will develop
concentration bounds which illustrate their interactions
throughout the rest of the paper.

\subsection{Relaxing the sparsity assumptions}
\label{sec::Uclass}
While the rates in Theorem~\ref{thm::main-II} are essentially tight,
we can tighten those in Theorem~\ref{thm::main} under an alternative
set of sparsity conditions.
In particular, relaxation of (A1) is feasible when we consider a
restricted uniformity class of inverse covariance matrices whose matrix
$\ell_1$ norm
is bounded by a parameter $M$: for $0\le q <1$,
\begin{eqnarray*}
&&\U_q\bigl(d_0(m), M\bigr)
\\
& &\qquad = \Biggl\{\Theta= (\theta_{ij})_{1 \le i, j \le m} : \max
_{i} \sum_{j=1}^m\vert
\theta_{ij}\vert^q \leq d_0(m), \llVert \Theta
\rrVert _1 \leq M, \Theta\succ0 \Biggr\}.
\end{eqnarray*}
It is to be understood that $d_0(m), M$ are positive numbers
bounded away from $0$ which are allowed to grow with $m, f, n$.
We focus on the case when $q = 0$ and consider positive definite matrices
with row/column sparsity constraints, upon which
we obtain a more refined result on the $\ell_2$ error bounds in
Theorem~\ref{thm::main-clime}.
First, we replace (A1) with (A1$'$), where $\Theta_0 = \rho(A_0)^{-1}$
and $\Phi_0 = \rho(B_0)^{-1}$ denote the inverse correlation matrices.
\begin{longlist}[(A1$'$)]
\item[(A1$'$)]
\label{assump1}
Suppose that $\Theta_0 \in\U_0(d_0(m), M)$ and
$\Phi_0 \in\U_0(d_0(f), K)$,
where $d_0(m)$, $d_0(f)$, $M$ and $K$ are positive and bounded away
from 0.
The dimensions $f$ and $m$ are allowed to grow with respect to each
other while the number of nonzero elements in each row or column of
$\Theta_0$ and $\Phi_0$ must be bounded by $d_0(m)$ and $d_0(f)$,
respectively: as $f, m \to\infty$
\[
\label{eq::sparA-clime} d_0(m) \llVert \Theta_0\rrVert
_1^2 = o \biggl(\frac{\sqrt {nf}}{\log^{1/2}
(m \vee f)} \biggr)\quad \mbox{and}\quad
d_0(f) \llVert \Phi_0\rrVert _1^2
= o \biggl(\frac{\sqrt {nm}}{\log^{1/2} (m
\vee f)} \biggr).
\]
\end{longlist}
We present Theorem~\ref{thm::main-clime}
using the CLIME estimators~\cite{CLL11}, which are obtained by first
solving the following
optimization functions:
%
\begin{eqnarray}
\label{eq::LM}  \tilde{\Theta} &=& \argmin_{\Theta\in\R^{m \times m}} \llvert \Theta\rrvert
_1 \qquad\mbox{subject to } \bigl\llVert \hat{\Gamma}(A_0)
\Theta- I\bigr\rrVert _{\max} \leq\lambda _{M},
\\
\label{eq::LK}  \tilde{\Phi} &= &\argmin_{\Phi\in\R^{f \times f}} \llvert \Phi \rrvert
_1\qquad \mbox{subject t o} \bigl\llVert \hat{\Gamma}(B_0)
\Phi- I\bigr\rrVert _{\max} \leq\lambda_{K}
\end{eqnarray}
for $\lambda_{M}$ and $\lambda_{K}$ to be specified in
Theorem~\ref{thm::main-clime};\vspace*{1pt}
then a symmetrization step selects each entry for the estimators
$\hat{\Theta}^{\clime}= (\hat\theta_{ij})$ and $\hat{\Phi
}^{\clime} = (\hat\phi_{ij})$,
as follows:
%
\begin{eqnarray}
\label{eq::A-Penest-corr-clime}  \hat{\Theta}_{\clime} &=&(\hat\theta_{ij})
\qquad\mbox{s.t. } \hat\theta_{ij} = \tilde\theta_{ij}\ind\bigl(|\tilde
\theta_{ij}| \leq |\tilde\theta_{ji}|\bigr) + \tilde
\theta_{ji}\ind\bigl(\vert\tilde\theta_{ij}\vert > \vert\tilde
\theta_{ji}\vert\bigr),
\\
\label{eq::B-Penest-corr-clime}  \hat{\Phi}_{\clime} &=& (\hat\phi_{ij}) \qquad
\mbox{s.t. } \hat\phi_{ij} =\tilde\phi_{ij}\ind\bigl(\vert\tilde
\phi_{ij}\vert \leq \vert\tilde\phi_{ji}\vert\bigr) + \tilde
\phi_{ji}\ind\bigl(\vert\tilde \phi _{ij}\vert > \vert\tilde
\phi_{ji}\vert\bigr).
\end{eqnarray}

\begin{theorem}
\label{thm::main-clime}
Consider data generating random matrices as in (\ref{data}).
Suppose that \textup{(A1$'$)} and \textup{(A2)} hold.
Let $\lambda_{A_0}$, $\lambda_{B_0}$ be as in Theorem~\ref
{thm::main}, and
\[
\lambda_{M} \asymp \llVert \Theta_0\rrVert
_1 \lambda_{B_0} \quad\mbox{and}\quad \lambda_{K} \asymp
\llVert \Phi_0\rrVert _1 \lambda_{A_0}\vadjust{\goodbreak}
\]
for $\lambda_{M}$, $\lambda_{K}$ as in~\eqref{eq::LM}
and~\eqref{eq::LK}. Let $\hat\Theta_{\clime}$ and $\hat\Phi
_{\clime}$ be as
in \eqref{eq::A-Penest-corr-clime}  and
\eqref{eq::B-Penest-corr-clime}.

Then with $\mathbb{P} (\X_0 ) \geq1- \frac{3}{(m \vee f)^2}$,
$\hat\Theta_{\clime}$ and $\hat\Phi_{\clime}$
are positive definite; and for $\hat{A \otimes B}$ as defined
in~\eqref{eq::ABestimate},
where $\hat{A}_{\rho} := \hat\Theta_{\clime}^{-1}$ and
$\hat{B}_{\rho} := \hat\Phi_{\clime}^{-1}$,
\begin{eqnarray}
\llVert \hat{A \otimes B} - A_0 \otimes B_0\rrVert
_2 &\leq&\llVert A_0\rrVert _2\llVert
B_0\rrVert _2 \delta \quad\mbox{and }\nonumber
\\
 \bigl\llVert \hat{A\otimes B}^{-1} - A_0^{-1}
\otimes B_0^{-1}\bigr\rrVert _2& \leq&\bigl
\llVert B_0^{-1}\bigr\rrVert _2 \bigl\llVert
A_0^{-1}\bigr\rrVert _2 \delta',\nonumber\\
\eqntext{\mbox{where }\displaystyle
 \delta, \delta' =O \bigl( \lambda_{A_0}
d_0(f)\llVert \Phi_0\rrVert ^2_1+
\lambda_{B_0} d_0(m)\llVert \Theta_0\rrVert
^2_1 \bigr) = o(1).}
\end{eqnarray}
\end{theorem}
Proof of Theorem~\ref{thm::main-clime} appears in
Section~16.

\begin{rems*}
Suppose that $f, m$ are sufficiently large, and
$f < m$. We focus our discussions on $\Theta_0$.
Denote the maximum node degree by $\degree(\Theta_0) := \max_{i}
\sum_{j}\ind(\theta_{ij} \neq0)$.
We note that (A1$'$) imposes the bounded node degree constraint in that:
$\degree(\Theta_0) = o (\sqrt{nf}/\log^{1/2} (m \vee f) )$,
while in (A1) a hub node alone can have up to $o(nf/\log(m \vee f))$
adjacent nodes.
Suppose that (A2) holds, and $n < \log(m \vee f)$.
In this case, (A1$'$) relaxes (A1) in the sense that it allows
$\degree(\Theta_0) = \Omega(1)$, and hence
$\llvert A_0^{-1}\rrvert _{0,\off} = \Omega(m)$, while (A1) does not.
Thus, the graphs considered in (A1) can be those which contain a single
or multiple disjoint components with some singleton nodes, while
those in (A1$'$) are allowed to be fully connected graphs.

Theorem~\ref{thm::main-clime} improves upon Theorem~\ref{thm::main} when
$M, K$ are slowly growing with respect to $m, f, n$, while
$d_0(m)$ and $d_0(f)$ are of lower order relative to
the total number edges in each graph.
However, this improvement requires that
we replace~\eqref{eq::A-Penest-corr-intro} and
\eqref{eq::B-Penest-corr-intro} with the CLIME or graphical
Dantzig-type estimators~\cite{CLL11,Yuan10}, for which we are able to
obtain faster rates of
convergence in the operator norm under (A1$'$) and (A2)
in estimating each covariance/correlation matrix.
The replacement is due to the lack of convergence bounds
on the $\ell_2$ errors which are tighter than those presented in
Theorem~\ref{thm::frob},
for the graphical Lasso estimators; as a consequence, the two summands
in $\delta$ and $\delta'$ in
Theorem~\ref{thm::main} were obtained using the rates of convergence
in the Frobenius norm, rather than the operator norm as we do in
Theorem~\ref{thm::main-clime}, for estimating the general (but sparse)
inverse correlation matrices.
To the best of our knowledge, comparable convergence bounds on the
operator norm for the graphical Lasso-type estimators are available
only under an \textit{irrepresentability condition} as developed in~\cite{RWRY08}.
We can indeed invoke their results in the present setting to
relax the sparsity constraint on in (A1), and
to prove faster rates of convergence in the operator norm
in view of Theorem~\ref{thm::large-devi-cor-multiple}.
\end{rems*}

\subsection{Related work}
\label{sec::related}
Algorithmic and theoretical properties of the graphical Lasso or Lasso-type
estimators have been well studied in the Gaussian graphical model
setting; see, for
example, \cite{MB06,YL07,LF07,FHT07,BGA08,RBLZ08,ZLW08,RWRY08,FFW09,PWZZ09}.\vadjust{\goodbreak}
Under sparsity and neighborhood stability conditions, the work
by~\cite{MB06} showed that the graph with $p$ nodes can be estimated
efficiently using the nodewise penalized regression approach using a
very small sample size $n$ in comparison
to the maximum node degree and the ambient dimension $p$.
The work of~\cite{Yuan10,CLL11,ZRXB11}, using variants of the approach
in~\cite{MB06}, showed convergence rates in the operator and the
Frobenius norm in estimating the precision matrix in case $p > n$,
where independent samples are always assumed.
It will be interesting to consider replacing the $\ell_1$ penalties
with the SCAD-type penalties or using the adaptive Lasso-type
penalties as in~\cite{LF07,FFW09}.
These approaches will reduce certain bias in the penalized estimators;
see, for example,
discussions in~\cite{FL01,zouli08}.
The recent work of~\cite{AT10} focuses on missing value imputation,
rather than estimation of the graphs or the underlying parameters.
When $f, m$ diverge as $n \to\infty$, the rates in~\cite{YL12} are
significantly slower than the corresponding ones in the present work.
Following essentially the same methods as in~\cite{AT10},
the same convergence rate as in \eqref{eq::frob-rate} on estimating
the covariance $\Sigma= A_0 \otimes B_0 $ in the Frobenius norm is obtained
in~\cite{LT12,THZ13}, in case $\llvert A_0^{-1}\rrvert _{0,\off}
=O(m)$, and
$\llvert B_0^{-1}\rrvert _{0,\off} = O(f)$; however,
this rate is obtained with the additional requirement that the number
of replicates
of $X$ must be at least on the order of
$n \geq\Omega ( (\frac{f}{m} \vee\frac{m}{f} )\log
\max(f,m,n) )$.
These results exclude the case for $n=1$ or for $n <\log(m \vee f)$,
which is the main focus of the present paper.
\section{Estimation of the correlation coefficients}
\label{sec::gemini-est}
In this section, we elaborate on two key technical results, namely,
the concentration bounds for sample correlation estimates and the
convergence bounds for the penalized inverse correlation estimates.

\subsection{Concentration bounds for sample correlations}
\label{sec::one-cent}
We now show the concentration bounds
for estimating the parameters in $\rho(A_0)$ and $\rho(B_0)$.
Theorem~\ref{thm::large-devi-cor-multiple} covers the small sample
settings, where the number of replications $n$ are
upper bounded by $\log(m \vee f)$, where
$m \vee f := \max(m, f)$.
We believe these are the first of such results to the best of
our knowledge. For completeness, we also state the bounds when $n >
\log(m \vee f)$ is large.

Let $K$ be the $\psi_2$ norm of $\xi$ for $\xi\sim\N(0, 1)$
defined as
%
\begin{eqnarray}
\label{eqdefineK}
&& K := \Vert \xi\Vert _{\psi_2} = \sup_{p \ge1} p^{-1/2}
 \bigl(\mathbb{E}\vert \xi\vert
^p\bigr)^{1/p};\qquad\mbox{thus, } K \mbox{ is the smallest }
K_2,\\
&&\qquad\mbox{which satisfies } \bigl(\mathbb{E}\vert \xi\vert ^p
\bigr)^{1/p }\le K_2\sqrt{p}\qquad \forall p \ge1; \mbox{see~\cite{vers12}}.
\end{eqnarray}

\begin{theorem}
\label{thm::large-devi-cor-multiple}
Consider data generating random matrices as in (\ref{data}).
Let $C$ be some absolute constant to be defined in~(57),
%
\begin{eqnarray}
\label{eq::tau0}  \tau_0& =& 2 C K^2 {
\log^{1/2}(m \vee f)}/{\sqrt{n}}\qquad \mbox{where $K$ is defined as in~\eqref{eqdefineK}},
\nonumber
\\[-8pt]
\\[-8pt]
\nonumber
\alpha_n& :=& {\llVert A_0\rrVert
_F \tau_0}/{\tr(A_0)}\quad\mbox{and}\quad
\beta_n := { \llVert B_0\rrVert _F
\tau_0}/{\tr(B_0)}.
\end{eqnarray}
Let $m \vee f \geq2$.
Then with probability at least $1- \frac{3}{(m \vee f)^2}$,
for $\alpha_n, \beta_n < 1/3$, and $\hat\Gamma(A_0)$ and $\hat
\Gamma(B_0)$
as in~\eqref{eq::sample-cor-Arep} and~\eqref{eq::sample-cor-Brep},
%
\begin{eqnarray*}
\forall i \neq j\qquad \bigl\llvert \hat\Gamma_{ij}(B_0)
-\rho _{ij}(B_0)\bigr\rrvert &\leq&\frac{\alpha_n}{1-\alpha_n} +
\bigl\llvert \rho_{ij}(B_0)\bigr\rrvert \frac{\alpha_n}{1-\alpha_n}
\leq3 \alpha_n,
\\
\forall i \neq j\qquad\bigl\llvert \hat\Gamma_{ij}(A_0)
-\rho_{ij}(A_0)\bigr\rrvert &\leq&\frac{\beta_n}{1-\beta_n} +\bigl
\llvert \rho_{ij}(A_0)\bigr\rrvert \frac{\beta_n}{1-\beta_n} \leq3
\beta_n
\end{eqnarray*}
and
\begin{equation}
\label{eq::zero-wei-bound} \Biggl\llvert \frac{1}{n}\sum
_{t=1}^n\bigl\llVert X(t)\bigr\rrVert
_F^2 - \tr(A_0) \tr(B_0)
\Biggr\rrvert \leq \tr(A_0) \tr(B_0) (
\alpha_n \wedge\beta_n).
\end{equation}
\end{theorem}

\begin{remark}
\label{rem::CA}
Note that under \textup{(A1)} and \textup{(A2)}, we have $\alpha_n, \beta_n \to0$
as $m$, $f \to\infty$, where
%
\begin{eqnarray}
\label{eq::alpha-rate}  \alpha_n \asymp C_A
\frac{\log^{1/2} (m \vee f)}{\sqrt{mn}} \quad\mbox{ and }\quad \beta_n \asymp C_B
\frac{\log^{1/2} (m \vee f)}{\sqrt{fn}} \nonumber\\[6pt]
\\[-26pt]
\eqntext{\mbox{where }\displaystyle C_A := \frac{\sqrt{m}\llVert  A_0\rrVert _F }{\tr(A_0)} = \frac{\sqrt{m}\sqrt{\tr(A_0
A_0)}}{\tr(A_0)}}\\
 \eqntext{\mbox{and }\displaystyle C_B := \frac{\sqrt{f}\llVert  B_0\rrVert _F}{\tr(B_0)} =\frac{\sqrt{f}\sqrt{\tr(B_0B_0)}}{\tr(B_0)}}
\end{eqnarray}
are treated to be constants throughout this paper under the bounded
spectrum assumptions in (A2).
Their magnitudes reflect how eigenvalues of each component
covariance matrix vary across its entire spectrum, and how much they
affect the estimation of the other matrix.
\end{remark}
The penalty parameters in Theorem~\ref{thm::main}
and~\ref{thm::main-II}, are chosen to dominate the
dominate the maximum of
entry-wise errors for estimating $\rho(A)$ and $\rho(B)$
with $\hat{\Gamma}(A)$ and $\hat{\Gamma}(B)$ as characterized in
Theorem~\ref{thm::large-devi-cor-multiple}:
\[
\label{eq::basic-penalty} \lambda_{A_0} \asymp C_A 
{
\log^{1/2} (m \vee f)}/{\sqrt{mn}} \quad\mbox{and}\quad \lambda_{B_0}
\asymp C_B {\log^{1/2} (m \vee f)}/{\sqrt{fn}}. 
\]
The notation $\lambda_{A_0}$ and $\lambda_{B_0}$ thus reflect their
dependencies on the eigenspectrum of $A_0$ and $B_0$,
which in turn affects the rate of convergence in the Frobenius norm
in estimating $\rho(B_0)$ and $\rho(A_0)$ with the penalized estimators.
The following large deviation bounds in
Lemma~\ref{lemma::large-devi-rep} are the key results in proving
Theorem~\ref{thm::large-devi-cor-multiple}.
We write it explicitly to denote by $\X_0$ the event that all large
deviation inequalities as stated in Lemma~\ref{lemma::large-devi-rep} hold.
%
\begin{lemma}
\label{lemma::large-devi-rep}
Suppose that \textup{(A2)} holds.
Denote by $\X_0$ the event that the following
inequalities hold simultaneously
for $\alpha_n, \beta_n$ as defined in~\eqref{eq::tau0}
\begin{eqnarray*}
\forall i, j\qquad \Biggl\llvert \frac{1}{n}\sum_{t=1}^n
\bigl\langle{ y(t)^i, y(t)^j }\bigr\rangle \Big/{\bigl(
\tr(A_0)\sqrt{ b_{ii} b_{jj}}\bigr)} -
\rho_{ij}(B_0)\Biggr\rrvert & \leq& \alpha_n,
\\
\forall i, j\qquad  \Biggl\llvert \frac{1}{n}\sum_{t=1}^n
\bigl\langle{ x(t)^i, x(t)^j }\bigr\rangle \Big/{\bigl(\tr
(B_0)\sqrt{a_{ii} a_{jj}}\bigr)} - \rho
_{ij}(A_0)\Biggr\rrvert & \leq& \beta_n.
\end{eqnarray*}
Suppose $m \vee f \geq2$.
Then $\mathbb{P} (\X_0 ) \geq1- \frac{3}{(m \vee f)^2}$.
\end{lemma}
The proofs for Theorem~\ref{thm::large-devi-cor-multiple}
and Lemma~\ref{lemma::large-devi-rep} appear in
Section~13.
We restate the first two inequalities Theorem~\ref
{thm::large-devi-cor-multiple}
in case $n=1$ in Theorem~\ref{thm::large-devi-cor}.
Let $C_A, C_B$ be as in Remark~\ref{rem::CA}.
%
\begin{theorem}
\label{thm::large-devi-cor}
Suppose $m \vee f \geq2$ and \textup{(A2)} holds.
Let $\hat\Gamma(A_0)$ and $\hat\Gamma(B_0)$ be as in~\eqref
{eq::sample-corr-A-intro}.
Then with probability at least $1- \frac{3}{ (m \vee f)^2}$, for all
$i \neq j$
\begin{eqnarray*}
 \bigl\llvert \hat\Gamma_{ij}(B_0) -
\rho_{ij}(B_0)\bigr\rrvert &\leq&2 C K^2
C_A \bigl(1 + \bigl\llvert \rho_{ij}(B_0)
\bigr\rrvert \bigr) \frac{ \log
^{1/2}(m \vee f)}{\sqrt{m}} \bigl(1+o(1)\bigr),
\\
 \bigl\llvert \hat\Gamma_{ij}(A_0) -
\rho_{ij}(A_0)\bigr\rrvert &\leq&2 C K^2
C_B \bigl(1 + \bigl\llvert \rho_{ij}(A_0)
\bigr\rrvert \bigr) \frac{ \log
^{1/2}(m \vee
f)}{\sqrt{f}} \bigl(1+o(1)\bigr).
\end{eqnarray*}
\end{theorem}

\begin{rems*}
We next compare the concentration bounds for the matrix normal
distribution as in Theorems~\ref{thm::large-devi-cor}
and~\ref{thm::large-devi-cor-multiple} with those for the multivariate
Gaussian.
First suppose that $f \le m$ and $B_0$ is an $f \times f$ identity matrix.
That is, we are given independent rows in $X$.
Then the rate of convergence for estimating $\rho_{ij}(A_0)$
with~\eqref{eq::sample-corr-A-intro} is bounded in~\cite{ZLW08,ZRXB11}
(cf. Lemma~13 and equation (43) in~\cite{ZRXB11}) as follows:
With probability at least $1 - 1/(f \vee m)^2$,
%
\begin{equation}
\label{eq::ratecomp} \bigl\llVert \hat{\Gamma}(A_0) -
\rho(A_0)\bigr\rrVert _{\max} < 3 C_3 \sqrt{{
\log(m \vee f)}/{f}}
\end{equation}
for $f$ large enough, so long as $m < e^{f/4C_3^2}$ for some constant
$C_3>4 \sqrt{5/3}$.
Now suppose that $B_0$ follows an $\operatorname{AR}(1)$ model with parameter $\rho$, then
the RHS of \eqref{eq::ratecomp} is necessarily replaced with a slower
rate of
%
\begin{equation}
\label{eq::rategem} \beta\asymp C_B {\log^{1/2} (m \vee f)}/{
\sqrt{f}}.
\end{equation}
We note that this rate as well as the rate of $\beta_n \asymp C_B \log
^{1/2} (m \vee f)/\sqrt{nf}$
are at the same order as the classical rate of \eqref{eq::ratecomp} as
the effective sample size
for estimating $A_0$ is $n f$ (cf. Remark~11.3).
However, both $\beta$ and $\beta_n$ are affected by the measure of~$C_B$,
which will increases as the parameter $\rho$ increases; we
illustrate this behavior in
our numerical results in Section~\ref{sec::roc-sum}.
We are able to remove the dependency on $C_B$
in~\eqref{eq::rategem} in Section~\ref{sec::flip-flop} under
additional sparsity conditions.
\end{rems*}

\subsection{Bounds on estimating the inverse correlation matrices}
\label{sec::inverse-frob}
In this section, we show explicit nonasymptotic convergence rates in the
Frobenius norm for estimating $\rho(A_0)$, $\rho(B_0)$ and their
inverses in Theorem~\ref{thm::frob}.
In Section~14, we present in
Corollary~14.1 a bound on the off-diagonal vectorized
$\ell_1$ norm on the error matrices for estimating
$\Theta_0 = \rho(A_0)^{-1}$
and $\Phi_0 = \rho(B_0)^{-1}$, which may be of independent interests.\vadjust{\goodbreak}

We say that event $\T(A_0)$ holds for sample correlation matrix
$\hat\Gamma(A_0)$ for some parameter $\delta_{n,f} \to0$, if for
all $j$,
$\hat{\Gamma}_{jj}(A_0) = \rho_{jj}(A_0) = 1$ and
%
\begin{equation}
\label{eq::delta-nf} \max_{j,k, j\neq k}\bigl|\hat{\Gamma}_{jk}(A_0)
- \rho_{jk}(A_0)\bigr| \leq \delta_{n,f},
\end{equation}
and the event $\T(B_0)$ holds for sample correlation matrix
$\hat\Gamma(B_0)$ for some parameter $\delta_{n,m} \to0$, if for
all $j$,
$\hat{\Gamma}_{jj}(B_0) = \rho_{jj}(B_0) = 1$ and
%
\begin{equation}
\label{eq::delta-nm} \max_{j,k, j\neq k}\bigl|\hat{\Gamma}_{jk}(B_0)
- \rho_{jk}(B_0)\bigr| \leq \delta_{n,m}.
\end{equation}

\begin{theorem}
\label{thm::frob}
Suppose that \textup{(A2)} holds.
Let $\hat{A}_{\rho}$ and $\hat{B}_{\rho}$ be the unique minimizers
defined by
\textup{\eqref{eq::A-Penest-corr-intro}} and~\textup{\eqref{eq::B-Penest-corr-intro}} with
sample correlation matrices $\hat\Gamma(A_0)$ and $\hat\Gamma(B_0)$
as their input. Suppose that event $\T(A_0)$ holds for $\hat\Gamma(A_0)$
for some $\delta_{n,f}$ and event $\T(B_0)$ holds for
$\hat\Gamma(B_0)$ for some $\delta_{n,m}$, such that
%
\begin{eqnarray}\label{eq::lambda-choir}
&& \delta_{n,f} \sqrt{\bigl\llvert
A_0^{-1}\bigr\rrvert _{0,\off} \vee1} = o(1)
\quad\mbox{and}\quad \delta_{n,m} \sqrt{\bigl\llvert B_0^{-1}
\bigr\rrvert _{0,\off} \vee 1}= o(1),
\nonumber
\\[-8pt]
\\[-8pt]
\nonumber
&&\qquad \mbox{set for some } 0< \e, \ve< 1, \lambda_{B} =
{\delta_{n,f}}/{\ve} \mbox{ and } \lambda_{A} = {
\delta_{n,m}}/{\e}.
\end{eqnarray}
Then on event $\T(A_0) \cap\T(B_0)$, we have for $9< C < 18$
\begin{eqnarray*}
\bigl\llVert \hat{A}_{\rho} - \rho(A_0)\bigr
\rrVert _2 \le \bigl\llVert \hat{A}_{\rho} -
\rho(A_0)\bigr\rrVert _F &\leq &C \Cond\bigl(
\rho(A_0)\bigr)^2 \lambda_{B} \sqrt{
\bigl\llvert A_0^{-1}\bigr\rrvert _{0,\off} \vee1},
\\
\bigl\llVert \hat{B}_{\rho} - \rho(B_0)\bigr
\rrVert _2 \le\bigl\llVert \hat{B}_{\rho} -
\rho(B_0)\bigr\rrVert _F &\leq& C \Cond\bigl(
\rho(B_0)\bigr)^2 \lambda_{A}\sqrt{
\bigl\llvert B_0^{-1}\bigr\rrvert _{0,\off} \vee1}
\end{eqnarray*}
and
%
\begin{eqnarray}
 \label{eq::eventAop} \bigl\llVert \hat{A}_{\rho}^{-1} -
\rho(A_0)^{-1}\bigr\rrVert _2 &\le&\bigl\llVert
\hat{A}_{\rho}^{-1} - \rho(A_0)^{-1}\bigr
\rrVert _F < \frac{C \lambda_{B} \sqrt{\llvert A_0^{-1}\rrvert _{0,\off} \vee
1}}{2\vp
^2_{\min}(\rho(A_0))},
\\
\label{eq::eventBop} \bigl\llVert \hat{B}_{\rho}^{-1} -
\rho(B_0)^{-1}\bigr\rrVert _2 &\le& \bigl\llVert
\hat{B}_{\rho}^{-1} - \rho(B_0)^{-1}\bigr
\rrVert _F \leq \frac{C \lambda_{A} \sqrt{\llvert B_0^{-1}\rrvert _{0,\off} \vee
1}}{2\vp^2_{\min
}(\rho(B_0))}.
\end{eqnarray}
\end{theorem}
Variants of Theorem~\ref{thm::frob} was shown in~\cite{RBLZ08}
in the context of Gaussian graphical models; our proof follows similar
arguments, and hence is omitted.
Lemma~\ref{lemma::eventA0} justifies the choices of the penalty
parameters $\lambda_{A_0}$ and $\lambda_{B_0}$.
%
\begin{lemma}
\label{lemma::eventA0}
Let $\alpha_n, \beta_n < 1/3$ be as defined in
Theorem~\ref{thm::large-devi-cor-multiple}.
Let
\begin{eqnarray*}
\label{eq::delta-nm}
 \delta_{n,f} &=& \frac{2\beta_n}{1-\beta_n} =O \biggl(
C_B \frac{\log^{1/2} (m \vee f)}{\sqrt{nf}} \biggr) \quad\mbox{and}\\
 \delta_{n,m} &=&
\frac{2 \alpha_n}{1-\alpha_n} =O \biggl(C_A \frac{\log^{1/2} (m \vee f)}{\sqrt{nm}} \biggr).
\end{eqnarray*}
Then, event $\T(A_0) \cap\T(B_0)$ holds on $\X_0$ for
the sample correlation matrices as defined in~\textup{\eqref{eq::sample-cor-Arep}}
and~\textup{\eqref{eq::sample-cor-Brep}}, respectively.

By Theorem~\ref{thm::large-devi-cor-multiple}, we have
$\mathbb{P} (\T(A_0) \cap\T(B_0) ) \geq1 - \frac{3}{(m
\vee f)^2}$.
\end{lemma}
%
\section{Variations on a theme}
\label{sec::MLE}
It is curious whether or not one can improve upon the Gemini
sample covariance/correlation estimators using the Flip-Flop
methods.
Essentially the Flip-Flop methods~\cite{Dut99,LZ05,WJS08} couple the
estimation for $A_0$ and $B_0$ by feeding the current estimate for
either of the two
into the likelihood function (or the penalized variants to be defined)
in order to optimize it with respect to the other.
Upon initialization of $A$ in \eqref{eq::alternate-A} with an identity
matrix, they obtain the
MLE for $A_0$ and $B_0$ by solving the following two equations
alternately and iteratively:
%
\begin{equation}
\label{eq::alternate-A} \quad  \tilde{B}(A) = \frac{1}{nm} \sum
_{t=1}^n X(t) A^{-1} X(t)^T,\qquad
\tilde{A}(B) = \frac{1}{nf} \sum_{t=1}^n
X(t)^T {B}^{-1} X(t)
\end{equation}
such that the corresponding output $\tilde{B}$, or $\tilde{A}$
becomes the
input as $B$, or $A$ to the RHS of equations in~\eqref{eq::alternate-A};
this process repeats until certain convergence criteria are reached.
The baseline Gemini method, where we simultaneously optimize a pair of
convex functions
\eqref{eq::A-Penest-corr-intro} and \eqref{eq::B-Penest-corr-intro},
can be seen as a single-step approximation of a penalized version of
\eqref{eq::alternate-A}, where we simply set both
$B$ and $A$ on the RHS of equations in~\eqref{eq::alternate-A} to be
the identity matrix.

We now introduce a natural variation of the Gemini estimators as given by
the Noniterative Penalized Flip-Flop (NiPFF) algorithm, where
we construct more sophisticated covariance and correlation estimators
based on
the pair of functions in~\eqref{eq::alternate-A}.

\textit{Noniterative Penalized Flip-Flop algorithm}:
\begin{enumerate}
\item[1.]
Assume $f \leq m$.
Initialize $A_{\init} = I$. Compute $\hat{\Gamma}(B_0)$ based on
\eqref{eq::sample-cor-Brep} as before, and compute $\hat{B}_{\rho}$
using GLasso~\eqref{eq::B-Penest-corr-intro} with the penalty parameter
$\lambda_{A_0}$ to be chosen (cf. Lemma~\ref{lemma::tildeAcov}). Let
$B_1 = \hat{W}_2 \hat{B}_{\rho}\hat{W}_2/m$.
\item[2.]
Now compute the sample covariance $\tilde{A}(B_1)$
using~\eqref{eq::alternate-A} and the sample correlation matrix $\hat
{\Gamma}(A_0)$ with
%
\begin{equation}
\label{eq::A-MLEcorr}  \hat{\Gamma}(A_0) = \tilde{W}_1^{-1}
\tilde{A}(B_1) \tilde{W}_1^{-1} \qquad\mbox{where }
\tilde{W}_1 = \diag\bigl(\tilde{A}(B_1)
\bigr)^{1/2}.
\end{equation}
Obtain an estimate $\hat{A}_{\rho}(B_1)$ using
GLasso~\eqref{eq::A-Penest-corr-intro} with $\hat{\Gamma}(A_0)$ in
\eqref{eq::A-MLEcorr} as its input, where $\lambda_{B} = \lambda
_{B_1}$ is to be specified
(cf. Remark~\ref{rem::lambdaB1}).

Let $A_1 = \hat{A}_* = \tilde{W}_1 \hat{A}_{\rho}(B_1)\tilde{W}_1$.
\item[3.]
Compute sample covariance matrix $\tilde{B}(A_1)$
using~\eqref{eq::alternate-A}, and
the sample correlation matrix $\hat{\Gamma}(B_0)$ with
%
\begin{equation}
\label{eq::B-MLEcorr}  \hat\Gamma(B_0) = \tilde{W}_2^{-1}
\tilde{B}(A_1)\tilde{W}_2^{-1} \qquad\mbox{where }
\tilde{W}_2 := \diag\bigl(\tilde{B}(A_1)
\bigr)^{1/2}.
\end{equation}
Obtain an estimate $\hat{B}_{\rho}(A_1)$
using~\eqref{eq::B-Penest-corr-intro},
with $\hat{\Gamma}(B_0)$ in \eqref{eq::B-MLEcorr} as its input,
where $\lambda_A = \lambda_{A_1}$ is to be specified
(cf. Theorem~\ref{thm::tildeBcorr} and Remark~17.8).

Let $\hat{B}_* = \tilde{W}_2 \hat{B}_{\rho}(A_1)\tilde{W}_2$.
\end{enumerate}

\section{Analysis for the penalized Flip-Flop algorithm}
\label{sec::flip-flop}
We illustrate the interactions between the row-wise and column-wise
correlations and covariances via the large deviation bounds to be
described in this section.
To make our discussion concrete,
suppose we aim to estimate $A_* = (a_{*,ij}) = m A_0 /\tr(A_0)$ and
$B_* = (b_{*,ij}) = B_0 \tr(A_0)/m$ instead of $A_0$ and $B_0$.
Note that $A_*$ has been normalized to have $\tr(A_*) = m$ for
identifiability. Let
%
\begin{equation}
 \label{eq::large-devi-local} \lambda_{f,n} = 2 C K^2
\frac{\log^{1/2} (m\vee f)}{\sqrt{fn}} \quad\mbox{and}\quad \lambda_{m,n} = 2 C K^2
\frac{\log^{1/2} (m \vee f)}{\sqrt{mn}},
\end{equation}
where $C$ is as in \eqref{eq::tau0} and $K$ as in~\eqref{eqdefineK}.
In analyzing the Flip-Flop algorithm,
we make the following additional assumption.
\begin{longlist}
\item[(A3)]
The inverse correlation matrices
have bounded $\llvert \rho(A_0)^{-1}\rrvert _1$ and\break  $\llvert \rho
(B_0)^{-1}\rrvert _1$:
\[
\bigl\llvert \rho(A_0)^{-1}\bigr\rrvert _1
\asymp m\quad \mbox{and}\quad \bigl\llvert \rho(B_0)^{-1}\bigr
\rrvert _1 \asymp f.
\]
\end{longlist}
First, we bound the entry-wise errors for the sample covariance and
correlation matrices as defined in step~2.
We note that the conclusions of Lemma~\ref{lemma::tildeAcov} and
Theorem~\ref{thm::tildeAcorr} continue to hold even if $\ve$ is chosen
outside of the interval $(0, 2/3]$, so long it is bounded away from
$0$ and $1$.
%
\begin{lemma}
\label{lemma::tildeAcov}
Suppose $(m \vee f) = o(\exp(m \wedge f))$.
Suppose that \textup{(A1)}, \textup{(A2)} and \textup{(A3)} hold.
Let $\hat{B}_{\rho}$ and $B_1$ be obtained as in step~1,
where we choose
\[
\lambda_{A_0} = \frac{2\alpha}{\ve(1-\alpha)} \ge \frac{3\alpha}{1-\alpha}\qquad
\mbox{for } \alpha= C_A \lambda_{m,n} \mbox{ where } C_A = {
\llVert A_0\rrVert _F\sqrt{m}}/{\tr(A_0)}
\]
and $0 < \ve< 2/3$.
Then on event $\A_1$, for $\tilde{A}(B_1)$ as defined in~\eqref
{eq::alternate-A}
%
\begin{eqnarray}
\label{eq::tildeA-corr-entry} & & \bigl\llvert \bigl(\tilde{A}(B_1) - A_*
\bigr)_{ij}\bigr\rrvert \leq \sqrt{a_{*,ii} a_{*,jj}}
\lambda_{f,n} \bigl(1+ o(1)\bigr) + \llvert a_{*,ij}\rrvert
\tilde{\mu},
\\
\label{eq::def-tildemu} &&\qquad \mbox{where } \tilde{\mu} = \lambda_{A_0}\bigl
\llvert \hat{B}_{\rho}^{-1}\bigr\rrvert _{1,\off}/{f} +
\frac{\alpha
}{1-\alpha} \bigl\llvert \hat{B}_{\rho}^{-1}\bigr\rrvert
_1/{f} \le\mu
\\
 \label{eq::define-mu}&&\qquad \mbox{for } \mu= \lambda_{A_0} {\bigl\llvert
\rho(B_0)^{-1}\bigr\rrvert _{1,\off}}/{f} +
\frac{\alpha}{(1-\alpha)} \bigl\llvert \rho(B_0)^{-1}\bigr\rrvert
_1/{f} + o(\lambda _{A_0} ).
\end{eqnarray}
Moreover, we have for some constant $d \le8$,
$\mathbb{P} (\A_1 ) \geq1-\frac{d}{(m \vee f)^2}$.
\end{lemma}

\begin{theorem}
\label{thm::tildeAcorr}
Suppose all conditions in Lemma~\ref{lemma::tildeAcov} hold.
Let $\hat{\Gamma}(A_0)$ be as defined in~\eqref{eq::A-MLEcorr}.
Then\vadjust{\goodbreak} on event $\A_1$, for $\tilde\eta:= \lambda_{f,n} (1+o(1))+
\tilde{\mu}$, where $\tilde{\mu}$ is as defined in~\eqref{eq::def-tildemu},
$\forall i\neq j$
%
\begin{eqnarray}
\label{eq::tildeA-corr-max} && \bigl\llvert \hat\Gamma_{ij}(A_0)
- \rho_{ij}(A_0)\bigr\rrvert
\nonumber
\\[-8pt]
\\[-8pt]
\nonumber
&&\qquad\leq\bigl(1+o(1)\bigr)
\lambda_{f,n} \bigl(1+ \bigl\llvert \rho_{ij}(A_0)
\bigr\rrvert \bigr) + \frac{2 \llvert \rho_{ij}(A_0)\rrvert \tilde{\mu
}}{1-\tilde\eta}
\\
\label{eq::define-eta} &&\qquad  \leq\frac{2 \eta}{1-\eta} \qquad\mbox{where } \eta=
\lambda_{f,n} \bigl(1+ o(1)\bigr) + \mu \mbox{ for $\mu$ as in~\eqref{eq::define-mu}.}
\end{eqnarray}
\end{theorem}
%
\begin{remark}
\label{rem::lambdaB1}
On event $\A_1$, the random quantities $\tilde{\mu}$ and
$\tilde{\eta}$ are upper bounded by $\mu$~\eqref{eq::define-mu}
and $\eta$~\eqref{eq::define-eta}, respectively, which can be
rewritten as follows.

Define $C_f := \llvert \rho(B_0)^{-1}
\rrvert _1/ {f} +\frac{2}{\ve}\llvert
\rho(B_0)^{-1}\rrvert _{1,\off}/{f}$ so that
\[
\mu= \frac{\alpha}{(1-\alpha)} \bigl(C_f + o(1) \bigr)\quad \mbox{and}\quad
\eta= \biggl(\lambda_{f,n} + \frac{\alpha}{(1-\alpha)} C_f\biggr)
\bigl(1+o(1)\bigr),
\]
which suggests that we set the penalty in step~2 in the
order of $\eta$,
\[
\label{eq::defineCf}  \lambda_{B_1} \asymp{2 \eta}/{(1-\eta)} \asymp
\lambda_{f,n} + C_f {\alpha}/{(1-\alpha)} C_f
\asymp\lambda_{f,n} + \lambda_{m,n}.
\]
Clearly $C_f \asymp1$ under \textup{(A3)}. Indeed, throughout this paper, we assume
%
\begin{equation}
 \label{eq::B1-pen} \lambda_{B_1} = \frac{2 \tilde\eta}{\ve_1(1-\tilde\eta)}\qquad \mbox{where } 0<
\ve_1 <1.
\end{equation}
\end{remark}
We compute the rates of convergence in the operator and the Frobenius
norm for estimating $A_*$ with $\hat{A}_*$ in step~2 in
Corollary~17.2 in Section~17.1.
The rates we obtain in Corollary~17.2
correspond to exactly those in Corollary~10.1 for the
baseline Gemini estimator, with slightly better leading constants.

Next, we bound the entry-wise errors for the sample
correlation matrix as defined in step~3
in Theorem~\ref{thm::tildeBcorr}. The corresponding result for
sample covariance is stated in Lemma~17.5.
%
\begin{theorem}
\label{thm::tildeBcorr}
Suppose $m \vee f = o(\exp(m \wedge f))$.
Suppose that \textup{(A1), (A2)}, and \textup{(A3)} hold.
Let $\hat{\Gamma}(B_0)$ be as defined in~\eqref{eq::B-MLEcorr}.
Let $\zeta= \lambda_{m,n} (1+o(1)) + \xi$, where $\xi$ is as defined
in~\eqref{eq::define-xi}.
Then on event $\A_1 \cap\event_2$, $\forall i\neq j$,
%
\begin{eqnarray}
\label{eq::tildeB-corr-entryI}&& \bigl\llvert \hat\Gamma_{ij}(B_0)
- \rho_{ij}(B_0)\bigr\rrvert
\nonumber
\\[-8pt]
\\[-8pt]
\nonumber
&&\qquad \leq \frac{\lambda_{m,n}(1+o(1))}{1-\zeta}+
\bigl\llvert \rho _{ij}(B_0)\bigr\rrvert \frac{\zeta+ \xi}{1-\zeta}
\\
\label{eq::tildeB-corr-entryII} &&\qquad\le\bigl(\lambda_{m,n}\bigl(1+\bigl\llvert
\rho_{ij}(B_0)\bigr\rrvert \bigr) + 2\bigl\llvert
\rho_{ij}(B_0)\bigr\rrvert \xi\bigr) \bigl(1+o(1)\bigr)
\\
\label{eq::define-xi}& & \qquad\quad \mbox{for } \xi= \lambda_{B_1}\bigl\llvert
\rho(A_0)^{-1}\bigr\rrvert _{1,\off}/{m} +
\frac{\eta}{1-\eta} \bigl\llvert \rho(A_0)^{-1}\bigr\rrvert
_1/ {m}+ o(\lambda _{B_1}).
\end{eqnarray}
Moreover, we have for some constant $d \le10$,
$\mathbb{P} (\A_1 \cap\event_2 ) \geq1-\frac{d}{(m
\vee f)^2}$.\vadjust{\goodbreak}
\end{theorem}

\subsection{Discussion}
\label{sec::mle-discuss}
Throughout this discussion, the $O(\cdot)$ notation hides a constant no
larger than $1+o(1)$.
We first compare the bound in \eqref{eq::tildeA-corr-max}
with that of Theorem~\ref{thm::large-devi-cor-multiple}, where
on event $\X_0$, for $\hat\Gamma(A_0)$ as defined
in~\eqref{eq::sample-cor-Arep},
%
\begin{equation}
\label{eq::one-step-rate} \forall i \neq j \qquad\bigl\llvert \hat\Gamma_{ij}(A_0)
-\rho_{ij}(A_0)\bigr\rrvert = O \bigl(C_B
\lambda_{f,n} \bigl(1+ \bigl\llvert \rho_{ij}(A_0)
\bigr\rrvert \bigr) \bigr),
\end{equation}
where $C_B = \llVert  B_0\rrVert _F\sqrt{f}/{\tr(B_0)}$.
On the other hand, the influence of $\lambda_{A_0} \asymp
\frac{2\alpha}{1-\alpha}$ on the entry-wise error for estimating
$\rho_{ij}(A_0)$ in \eqref{eq::tildeA-corr-max}
is regulated through both $C_f$, which is a bounded constant under (A3)
(see Remark~\ref{rem::lambdaB1}),
as well as the magnitude of $\rho_{ij}(A_0)$ itself; to see this,
by \eqref{eq::tildeA-corr-max}, $\forall i \neq j$,
\[
\bigl\llvert \hat\Gamma_{ij}(A_0) - \rho_{ij}(A_0)
\bigr\rrvert =  O \bigl(\lambda_{f,n}\bigl(1+\bigl\llvert
\rho_{ij}(A_0)\bigr\rrvert \bigr) + 2 \bigl\llvert
\rho_{ij}(A_0)\bigr\rrvert C_f C_A
\lambda_{m,n} \bigr).
\]
This rate is in the same order as that in~\eqref{eq::one-step-rate}.
However, when $\lambda_{m,n} \ll\lambda_{f,n}$,
the second term is
of smaller order compared to the first term. In this case,
the upper bound in \eqref{eq::tildeA-corr-max} is dominated by the
first term on the RHS, and
one can perhaps obtain a slightly better bound with Theorem~\ref
{thm::tildeAcorr}, as
the leading term no longer depends on the constant $C_B$ as displayed
in \eqref{eq::one-step-rate}.

We next compare the bound in \eqref{eq::tildeB-corr-entryI}
with that of Theorem~\ref{thm::large-devi-cor-multiple}.
Before we proceed, we first define the following parameter:
\begin{eqnarray*}
 C_m &=& \bigl\llvert \rho(A_0)^{-1}\bigr
\rrvert _1/ {m}+\frac{2}{\ve_1}\bigl\llvert \rho(A_0)^{-1}
\bigr\rrvert _{1,\off}/{m}\qquad \mbox{so that}
\\
 \xi&\le&\frac{\eta}{1-\eta} \bigl(C_m+o(1)\bigr) \quad\mbox{and}\quad \zeta
\le\biggl( \lambda_{m,n} + \frac{\eta}{1-\eta} C_m\biggr)
\bigl(1+o(1)\bigr),
\end{eqnarray*}
where $0 < \ve_1 < 1$ is the same as in~\eqref{eq::B1-pen}.
Hence, we have on $\A_1 \cap\event_2$, by \eqref{eq::tildeB-corr-entryI},
\[
\nonumber
\bigl\llvert \hat\Gamma_{ij}(B_0) -
\rho_{ij}(B_0)\bigr\rrvert  =  O \biggl(
\lambda_{m,n} \bigl(1+ \bigl\llvert \rho_{ij}(B_0)
\bigr\rrvert \bigr) + \bigl\llvert \rho_{ij}(B_0)\bigr
\rrvert C_m \frac{2\eta}{1-\eta
} \biggr),
\]
where $2\eta/(1-\eta) \asymp\lambda_{m,n} + \lambda_{f,n}$.
Clearly, the influence of $\lambda_{B_1} \asymp\frac{2\eta}{1-\eta}$
on the entry-wise error for estimating $\rho_{ij}(B_0)$
is regulated through the quantity $C_m$ which is a constant
under (A3), as well as the magnitude of $\rho_{ij}(B_0)$.

We note that when $m \asymp f$,
these rates are in the same order of
$O (\lambda_{m,n} (1 + \llvert \rho_{ij}(B_0)\rrvert
) )$
as those in Theorem~\ref{thm::large-devi-cor-multiple} on event $\X_0$.
Moreover, for pairs of $(i, j)$ where $i \neq j$, such that
$\llvert \rho_{ij}(B_0)\rrvert $ is small, one can perhaps
obtain a
slightly better bound with Theorem~\ref{thm::tildeBcorr}, as
the first (leading) term which involves $\lambda_{m,n}$
no longer depends on the constant
$C_A \geq1$ as needed in~\eqref{eq::alpha-rate}.
In summary, for the following two cases, we expect that the sample
correlation estimate $\hat\Gamma(B_0)$ which we obtain in
step~3 improves upon the initial estimate in step~1:
\begin{longlist}[1.]
\item[1.]
For all $i \neq j$, $\rho_{ij}(B_0)$ is bounded in magnitudes;
for example, when $\llvert \rho_{ij}(B_0)\rrvert  =O( \sqrt{f/m})$,
then $\zeta' \asymp\lambda_{m,n}$.
In particular, for $\rho(B_0) = I$,
\[
\forall i\neq j \qquad\bigl\llvert \hat\Gamma_{ij}(B_0) - \rho
_{ij}(B_0)\bigr\rrvert \leq2\lambda_{m,n}
\bigl(1+o(1)\bigr).
\]
Hence, the error in estimating $A_0$ is propagated into
the estimate of $\rho_{ij}(B_0)$ only when $\rho_{ij}(B_0) \neq 0$.
\item[2.]
When $m$ and $f$ are close to each other
in that the ratio $m/f \to\const>1$, and simultaneously, $C_m$,
$C_f$, and $\llvert \rho_{ij}(B_0)\rrvert $ are small for all
$i \neq j$; then
$2\zeta' =2 (\lambda_{m,n}+ \max_{i\neq j} \llvert \rho
_{ij}(B_0)\rrvert \xi)
\asymp\lambda_{m,n} + \lambda_{f,n}$
provides a tight upper bound for the RHS of~\eqref{eq::tildeB-corr-entryII}.
\end{longlist}
Suppose that $m \gg f$.
Then the original estimator in~\eqref{eq::sample-cor-Brep}
could be much better for pairs of $(i,j)$
with a large $\llvert \rho_{ij}(B_0)\rrvert $. As for such
pairs, the second
term is of larger order than the first term in~\eqref{eq::tildeB-corr-entryII}.
A refined analysis on the GLasso given the estimates in
Theorem~\ref{thm::tildeBcorr} is left as future work.

\section{Numerical results}
\label{sec::numerical}
We demonstrate the effectiveness of the Gemini method as well as the
Noniterative Penalized Flip-Flop method, which we refer to as
the FF method, with simulated data.
We also show an example of applying Gemini to a real data set, the EEG data,
obtained from UCI Machine learning repository \cite{UCIeeg} in
Section~\ref{subsec::eeg}.
For a penalty parameter $\lambda\ge0$, the GLasso estimator is given by
\[
\label{logliki} \glasso(\hat\Gamma, \lambda) = \mathop{\operatorname
{arg\,min}}_{\Theta \succ0} \bigl(\mbox{tr}(\hat{\Gamma} \Theta) - \log|\Theta| +
\lambda\llvert \Theta\rrvert _{1,\off}\bigr),
\]
where $\hat\Gamma$ is a sample correlation matrix.
We use the R-package \texttt{glasso}~\cite{FHT07} to compute the
GLasso solution. For the two estimation methods, we have
various tuning parameters, namely $\lambda$, $\nu$ for the baseline
Gemini estimators,
and $\phi, \upsilon$ for the FF method.
In our simulation study, we look at three different models from which
$A$ and $B$ will be chosen. Let $\Omega= A^{-1} = (\omega_{ij})$ and
$\Pi= B^{-1} = (\pi_{ij})$. Let $E$ denote edges in $\Omega$, and
$F$ denote edges in $\Pi$.
We choose $A$ from one of these two models:
\begin{itemize}
\item $\operatorname{AR}(1)$ model.
In this model, the covariance matrix is of the form $A =\{\rho
^{|i-j|}\}_{i,j}$.
The graph corresponding to $\Omega$ is a chain.
\item Star-Block model.
In this model the covariance matrix is block-diagonal with equal-sized
blocks whose inverses correspond to star structured graphs, where
$A_{ii} = 1$, for all $i$.
We have 20 subgraphs, where in each subgraph, 8 nodes are
connected to a central hub node with no other connections.
The rest of the nodes in the graph are singletons.
Covariance matrix for each block $S$ in $A$ is generated as in \cite{RWRY08}:
$S_{ ij} = \rho= 0.5$ if $(i,j) \in E$ and $S_{ij} = \rho^2$
otherwise.
\end{itemize}
For $\Pi$, we use the random concentration matrix model in
\cite{ZLW08}. The graph is generated according to a type of Erd\H
{o}s--R\'{e}nyi random graph
model. Initially, we set $\Pi= 0.25 I_{f \times f}$, where $f = 80$.
Then we randomly select $d$ edges and update $\Pi$ as follows:
for each new edge $(i, j)$, a weight $w >0$ is chosen uniformly at
random from $[w_{\min}, w_{\max}]$ where $w_{\max} > w_{\min} > 0$;
we subtract $w$ from $\pi_{ij}$ and $\pi_{ji}$, and increase $\pi
_{ii}$ and $\pi_{jj}$ by $w$.\vadjust{\goodbreak} This keeps $\Pi$ positive definite.
For both models of $A$, we have $A_* = A \frac{m}{\tr(A)} = A = \rho(A)$.
Let $\Omega_* = \frac{\tr(A)}{m}\Omega$ and $\Pi_* =
\frac{m}{\tr(A)}\Pi$. Thus,\vspace*{1pt} we have $\Omega_* = \Omega$ and $\Pi
_* =
\Pi$ for all combinations of $A$ and $B$ in this section.

%
\begin{table}
\caption{Metrics for evaluating $\hat{E}(\lambda)$}
\label{tab:fpfn}
\begin{tabular*}{\textwidth}{@{\extracolsep{\fill}}ll@{}}
\hline
\multicolumn{1}{@{}l}{\textbf{Metric}} & \textbf{Definition} \\
\hline
False positives (FPs) & \# of incorrectly selected edges in $\hat
{E}(\lambda)$:
$|\hat{E}(\lambda) \setminus E|$ \\
False negatives (FNs) & \# of edges in $E$ that are not selected in
$\hat{E}(\lambda)$:
$|E \setminus\hat{E}(\lambda)|$ \\
True positives (TPs) & \# of correctly selected edges: $|\hat
{E}(\lambda) \cap E|$ \\
True negatives (TNs) & \# of zeros in $\hat{E}(\lambda)$ that are
also zero in $E$ \\
False positive rate ($\fpr$) & $\fpr= \mathrm{FP} / (\mathrm{FP} + \mathrm{TN}) = \mathrm{FP}/\bigl({m \choose
2} -\llvert E\rrvert \bigr) $ \\
False negative rate ($\fnr$) & $ \fnr= \mathrm{FN}/ (\mathrm{TP}+\mathrm{FN}) = \mathrm{FN} /|E|$ \\
MCC & $ \mathrm{TP} \times \mathrm{TN}/\sqrt{(\mathrm{TP}+\mathrm{FP})(\mathrm{TP}+\mathrm{FN})(\mathrm{TN}+\mathrm{FP})(\mathrm{TN}+\mathrm{FN})}$ \\
\hline
\end{tabular*}
\end{table}

\subsection{Regularization paths and cross-validation}
\label{sec::paths}
We illustrate the behaviors of the Gemini estimators for each model
combination of $A, B$ with $m=400$ and $f=80$ over the full
regularization paths.
To evaluate consistency, we use relative errors in the operator and the
Frobenius norm.
For model selection consistency, we use false positive and false
negative rates and Matthews correlation coefficient (MCC) as defined in
Table~\ref{tab:fpfn}.
For each pair of covariance matrices, we do the following.
First, we generate $A$ and $B$, where $A$ is $m \times m$ and $B$ is $f
\times f$.
Let $A^{1/2}$ and $B^{1/2}$ be the unique square root of matrix $A$
and $B$, respectively. Let $T$ and $T'$ be a set of values in $(0, 0.5]$.
Now, repeat the following steps 100 times:
\begin{enumerate}
\item
Sample random matrices
$X^{(1)}, \ldots, X^{(n)} \mbox{ i.i.d. } \sim\N_{f,m} (0, A \otimes B)$:
\[
\label{eq::matrix-normal-rep} X^{(t)} = B^{1/2} Z(t) A^{1/2}
\qquad\mbox{where } Z_{ij}(t) \sim N(0, 1)\ \forall i, j, \forall t = 1, \ldots, n.
\]
Compute the sample column correlation $\hat\corr_{\mathrm{col}}$ as in \eqref
{eq::sample-cor-Arep}
and row correlation $\hat\corr_{\mathrm{row}}$ as in \eqref{eq::sample-cor-Brep}.
\item
For each $\lambda\in T$ and $\nu\in T'$:
\begin{enumerate}[(a)]
\item[(a)]
Obtain the estimated inverse correlation matrices
$\hat{A}^{-1}$, and $\hat{B}^{-1}$ with
$\glasso(\hat\corr_{\mathrm{col}}, \lambda)$ and $\glasso(\hat\corr
_{\mathrm{row}}, \nu)$, respectively.\vspace*{1pt}
Let $\hat\Omega(\lambda) := \hat{A}_*^{-1}$ and $\hat\Pi(\nu) :=
\hat{B}_*^{-1}$,
where $\hat{A}_*$ and $\hat{B}_*$ are as defined in~(42).
\item[(b)]
Let $\hat{E}(\lambda)$ denote the set of edges in the estimated $\hat
{\Omega}(\lambda)$.
Now compute $\fnr(\lambda)$, $\fpr(\lambda)$ and $\mcc(\lambda)$
as defined in Table~\ref{tab:fpfn}.
To obtain $\fnr(\nu)$, $\fpr(\nu)$ and $\mcc(\nu)$,
we replace $\hat{E}(\lambda)$ with $\hat{F}(\nu)$,
which denotes the set of edges in $\hat{\Pi}(\nu)$, $E$ with $F$,
and $m$ with $f$.
Compute the relative errors
$\|\hat{\Omega}(\lambda)-\Omega\|/\llVert \Omega\rrVert $ and
$\|\hat{\Pi}(\nu)-\Pi\|/\llVert \Pi\rrVert $,
where $\llVert \cdot\rrVert $ denotes the operator or the
Frobenius norm.
\end{enumerate}
\end{enumerate}

\begin{figure}

\includegraphics{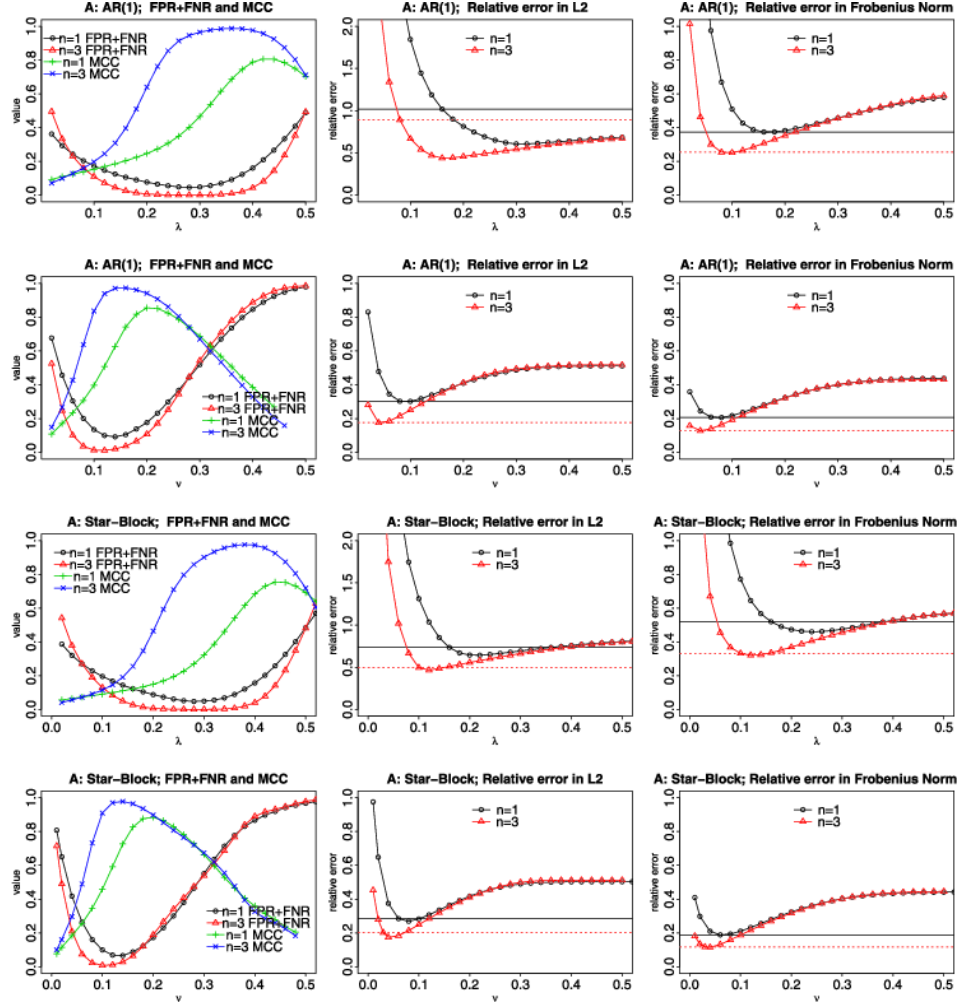}

\caption{$m=400$, $f=80$.
$B^{-1}$ follows random graph model with $d = 80$ and $w \in[0.1,
0.3]$
throughout these plots.
In the top two panels, covariance $A$ follows the $\operatorname{AR}(1)$ model with
$\rho= 0.5$; for the bottom two panels, $A$ follows the Star-Block
model.
The top and the third panel are for $\hat\Omega(\lambda)$;
the second and the bottom panel are for $\hat\Pi(\nu)$.
As $\lambda$ or $\nu$ increases, FPs decrease
and FNs increase. As a result, $\mathrm{FPR}+\mathrm{FNR}$ first
decreases and then increases,
and on the other hand, $\mathrm{MCC}$ first increases as the estimated
graphs becomes
more accurate, and then decreases due to missing edges caused by large
penalization.
The relative errors also first decrease and
then increase before leveling off. This is because
decreased FPs first help reduce the estimation errors;
however, as penalization increases, the estimated
graphs miss more and more edges until only diagonal entries remain
in the inverse covariance estimates.
Solid and dashed horizontal lines in the second and third columns
show the performances of Gemini for
cross-validated tuning parameters: in the top two panels,
$\lambda_{\mathrm{CV}} = 0.16$ and $\nu_{\mathrm{CV}} = 0.08$ for $n=1$,
and $\lambda'_{\mathrm{CV}} = 0.08$ and $\nu'_{\mathrm{CV}} = 0.04$ for $n=3$.
For the bottom two panels, $\lambda_{\mathrm{CV}} = 0.16$ and $\nu_{\mathrm{CV}} = 0.10$
for $n=1$, and $\lambda'_{\mathrm{CV}} = 0.06$ and $\nu'_{\mathrm{CV}} = 0.03$ for $n=3$.
These tuning parameters tend to stay near the
$\lambda$ or $\nu$ that minimizes the relative error in the Frobenius
norm.}
\label{fig:chain-rand}
\end{figure}

After 100 trials, we plot each of the following as $\lambda$
changes over a range of values in $T$: $(\overline\fnr+ \overline
\fpr)(\lambda)$ and $\overline\mcc(\lambda)$ for
$\hat{E}(\lambda)$, where $\overline\fnr$, $\overline\fpr$ and
$\overline\mcc$ are
averaged over the 100 trials, and the average relative
errors in the operator and the Frobenius norm. Similarly, we plot these
as $\nu$ changes over a range of values in $T'$.
Figure~\ref{fig:chain-rand} shows how these four metrics
change as the $\ell_1$ regularization parameters $\lambda$ and $\nu$
increase over full paths where covariance $A$ comes from either
$\operatorname{AR}(1)$ or the Star-Block model, and
$\Pi$ comes from the random graph model.
These plots show that the Gemini method is able to select the correct
structures as well as achieving low relative errors in the operator and
the Frobenius norm when $\lambda$ and $\nu$ are chosen from a
suitable range. In addition, as $n$ increases, we see performance
gains over almost the entire paths for all metrics as expected.
Other model combinations of $A, B$ which are not shown here confirm
similar findings.

In Figure~\ref{fig:chain-rand}, we also illustrate choosing the
penalty parameters $\lambda$ and $\nu$ by 10-fold cross-validation. To
do so, we run the following for 10 trials.
In each trial, we partition the rows of each $X^{(t)}, t=1, \ldots, n$
into 10 folds.
For each fold, the validation set consists of the subset of rows of
$X^{(1)}, \ldots, X^{(n)}$
sharing the same indices and its complement set serves as the training data.
Denote by $\hat\corr_{T}$ and $\hat\corr_{V}$
the column-wise sample correlations based upon the training and the
validation data,
which are computed in the same manner as in~\eqref{eq::sample-cor-Arep}.
We define $\score_A(\lambda) = \mbox{tr}(\hat{\Theta}_\lambda\hat
{\corr}_{V}) - \log|\hat{\Theta}_\lambda|$, where $\hat\Theta
_\lambda= \glasso(\hat\corr_{T}, \lambda)$.
The final score for a particular $\lambda$ is the average over
10 trials (with 10 folds in each trial) and the one with the lowest score
is chosen to be $\lambda_{\mathrm{CV}}$.
Similarly, we use column partitions to obtain $\nu_{\mathrm{CV}}$.
We leave the theoretical analysis on cross-validation as future work.
\subsection{ROC comparisons}

In this section, we compare the performances of the two methods,
namely, the baseline Gemini and its three-step FF variant over
the full paths by examining their ROC curves.
Each curve is an average over 50 trials.
We fix $f= 80$, $m = 400$, $n=1$.

\begin{figure}

\includegraphics{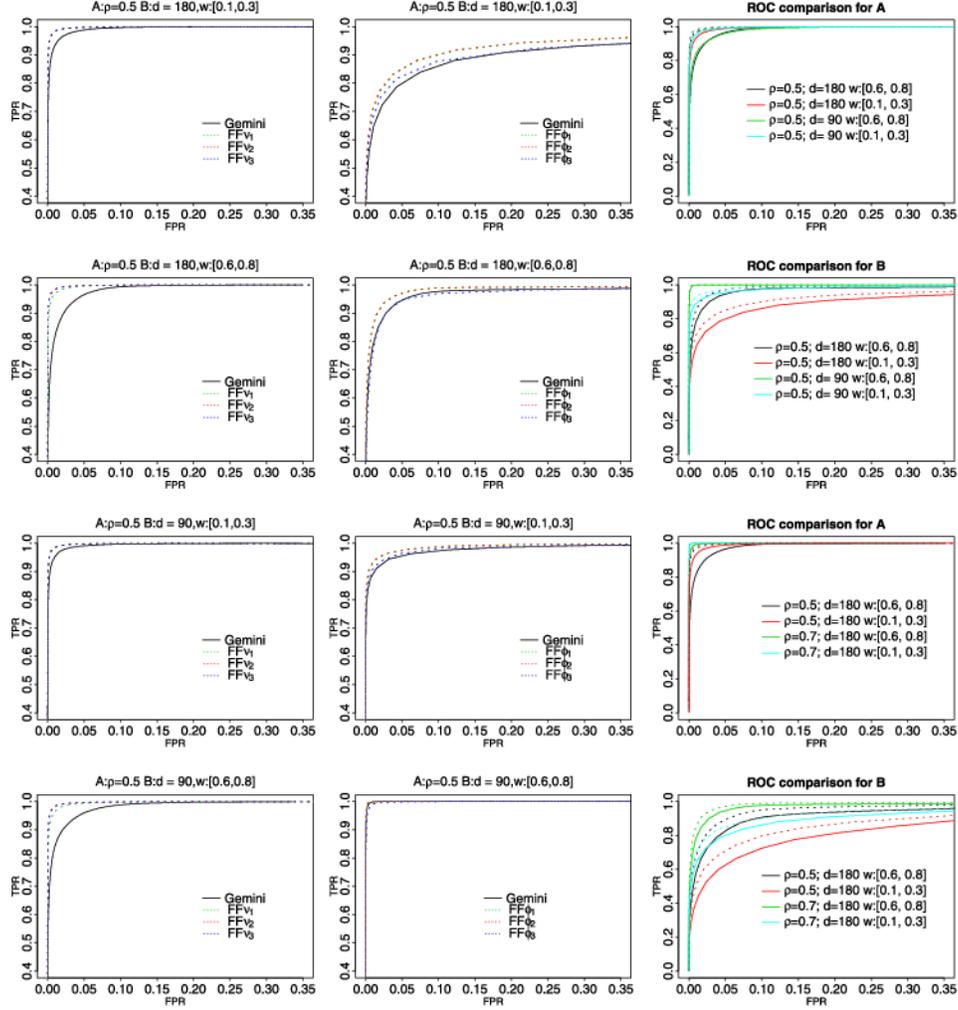}

\caption{$m=400$, $f=80$, $n=1$. Solid lines are for Gemini.
Plots in the left column are for $A$ and the middle column are for $B$.
The three dotted lines
in each plot on the left column correspond to the three optimization
criteria $\nu_1,
\nu_2, \nu_3$ as specified in step 3.
For the middle column, they correspond to $(i_1, \phi_1), (i_2,
\phi_2), (i_3, \phi_3)$, as specified in step 4. In the right column:
in top two plots, we choose $A$ from $\operatorname{AR}(1)$ model with $\rho= 0.5$
while changing the settings of $B^{-1}$ as in Table~\protect\ref{tab2};
in bottom two plots, we choose $A$ from $\operatorname{AR}(1)$ model with $\rho= 0.5$
or $0.7$ while changing the settings of $B^{-1}$ with $d = 180$.
Dotted lines in plots for $A$ on the right column are chosen according
to the optimization criterion $\nu_1$, and in plots for $B$, they
are chosen according to the criterion $\phi_1$.}
\label{fig:roc-AB}
\end{figure}

To simplify our notation, we summarize the penalty parameters which we use
for indexing the ROC curves as follows:
\[
\lambda= \lambda_{B_0}, \qquad\nu= \lambda_{A_0},\qquad \phi=
\lambda_{B_1},\qquad \upsilon= \lambda_{A_1}.
\]
To illustrate the overall performances
of the baseline Gemini method for estimating the graphs of $\Omega$
and $\Pi$,
we use pairs of metrics $ (\overline{\fpr}(\lambda), 1-
\overline{\fnr
}(\lambda) )$
and $ (\overline{\fpr}(\nu), 1- \overline{\fnr}(\nu)
)$, respectively,
which we obtain as the average over 50 trials of steps 1 and 2 as
described in Section~\ref{sec::paths}.
To plot the ROC curves for the FF method, we start with estimating
$\Pi$ with the Gemini estimator. Due to computational complexity,
we specify the input parameters of the subsequent steps sequentially.
These choices are not feasible in practical settings.
We run through this idealized example for
the sake of comparing with the baseline Gemini estimators.
Repeat the following 50 times: Let $T := \{0.02, 0.04, \ldots, 0.72\} $.
\begin{longlist}[1--2.]
\item[1--2.]
Run steps 1, 2 as in Section~\ref{sec::paths}, while only computing the
metrics for $\hat\Pi(\nu)$, where $\nu\in T$.
\item[3.]
To execute the second step of the FF algorithm, we use the following
three outputs
from step 2 of the current procedure to act as $B_1$ to compute $\tilde
{A}(B_1)$.
We choose the output $B_1$ such that its corresponding $\nu$ is
chosen to be $\nu_1 = \arg\min_{\nu\in T} (\fnr+ \fpr)(\nu)$,
$\nu_2 = \arg\min_{\nu\in T} \|\hat{\Pi}(\nu)-\Pi\|_2/\llVert \Pi\rrVert _2$,
and $\nu_3 = \arg\min_{\nu\in T} \|\hat{\Pi}(\nu)-\Pi\|_F/\llVert \Pi\rrVert _F$.
Denote these by $B_1^1, B_1^2$ and $B_1^3$.
We now run the second step of the FF method for each $B_1^i$,
where $i=1, 2, 3$, with penalty parameter $\phi\in T$ changing over
the full path
while obtaining the inverse estimators $\hat{\Omega}^i(\phi)$ for
$\Omega$ and computing $\fnr^i(\phi)$, and $\fpr^i(\phi)$ for each
estimated edge set.
These contribute to 3 ROC curves for estimating the edges in $E$.
\item[4.]
To execute the last step of the FF method, we use the following three outputs
from step 3 as $A_1$ to compute $\tilde{B}(A_1)$.
We choose the output $A_1$ such that its corresponding $(i,\phi)$ is
chosen to be
optimal with respect to one of the following metrics:
$(i_1, \phi_1) = \arg\min_{\phi\in T, i=1, 2, 3} (\fnr^i +
\fpr^i)(\phi)$, $(i_2, \phi_2) = \arg\min_{\phi\in T, i=1, 2, 3}
\|\hat{\Omega}^i(\phi) -\Omega\|_2/\llVert \Omega\rrVert
_2$, and $(i_3,
\phi_3) = \arg\min_{\phi\in T, i=1, 2, 3} \|\hat{\Omega}^i(\phi)
-\Omega\|_F/\llVert \Omega\rrVert _F$.
The choices then become $(\nu_{i_j}, \phi_j), j =1, 2, 3$, which we
simply denote by $\phi_1, \phi_2, \phi_3$.
Thus, there are again three choices for $A_1$.
We now run the third step of the FF method for each $\tilde{B}(A_1)$
with $\upsilon\in T$ changing over the full path,
while computing $\fnr^j(\upsilon)$ and $\fpr^j(\upsilon)$, where $j
=1, \ldots, 3$,
for each estimated edge set.
These contribute to 3 ROC curves for estimating the edges of $F$.
\end{longlist}
The ROC curves are plotted in Figure~\ref{fig:roc-AB} using pairs of metrics
$(\overline{\fpr}^i(\phi), 1- \overline{\fnr}^i(\phi))$ and
$(\overline{\fpr}^j(\upsilon), 1- \overline{\fnr}^j(\upsilon))$,
$i, j = 1,
2, 3$,
which are averaged over 50 trials.
Throughout the plots on the left
column of Figure~\ref{fig:roc-AB},
we see clear performance gains of the FF method over the
baseline Gemini on estimating $\Omega= A^{-1}$, when the initial
penalty $\nu$ is chosen properly. For $\Pi= B^{-1}$ in the middle column,
we do not always see improvements when $w$ is drawn from $[0.6,
0.8]$. We do see some improvements in case $w$ is drawn from $[0.1,
0.3]$ and when the total correlation $\rho_{B}^2$ is small. Overall,
the performance gains for $\Pi$ are not as substantial as those for
$\Omega$. These observations are consistent with our theory and
discussion in Section~\ref{sec::mle-discuss}.

\subsection{Summary on the ROC curves}
\label{sec::roc-sum}
We use the following metrics to compare matrix $B$ and $A$ across
different models or parameters:
\begin{longlist}[1.]
\item[1.]
Total correlation: $\rho_A^2 = \sum_{i < j} \rho_{ij}^2(A)/{m
\choose2}$ and
$\rho_B^2 = \sum_{i < j} \rho^2_{ij}(B)/{f \choose2}$.
\item[2.]
$\llVert  B\rrVert _F/\tr(B)$ and $\llVert  A\rrVert
_F/\tr(A)$: these affect
the entry-wise error bound in sample correlation estimates for
$\rho_{ij}(A)$ and $\rho_{ij}(B)$, for all $i\neq j$, for the
baseline Gemini estimators.
\item[3.]
The pairs of $\ell_1$-metrics
$(\llvert \rho(B)^{-1}\rrvert _{1,\off}, \llvert \rho(B)^{-1}\rrvert _1)$ and
$(\llvert \rho(A)^{-1}\rrvert _{1,\off},\break   \llvert \rho(A)^{-1}\rrvert
_1)$.\vadjust{\goodbreak}
\end{longlist}
The total correlation metric comes from~\cite{Efr09}.
We use it to characterize the average squared magnitudes for correlation
coefficients of $\rho(A)$ or $\rho(B)$. They are clearly relevant for
the FF method as the entry-wise error bound for estimating
$\rho_{ij}(A)$ and $\rho_{ij}(B)$, for all $i \neq j$,
depends on the magnitude of the entry itself
(cf. Theorems~\ref{thm::tildeAcorr} and~\ref{thm::tildeBcorr}).

We summarize our findings across the ROC curves in the right column in
Figure~\ref{fig:roc-AB}.
First, we focus on the case when $A$ is fixed and $B$ is changing.
When $\Pi$ follows the random graph model, we observe that
for both the baseline Gemini estimators and their FF variants,
the performances in terms of estimating edges for $\Omega$ are better
when the weights for $\Pi$ are chosen from $[0.1, 0.3]$ for both $d =
90$ and
$d=180$. Here, the sparsity for $\Pi$ is not the decisive factor.
This is consistent with our theory, in view of Table~\ref{tab2}, that $\llVert  B\rrVert _F/\tr(B)$ affects the
entry-wise error bound for the baseline Gemini correlation estimate
$\hat\Gamma(A)$ as shown in Theorem~\ref{thm::large-devi-cor},
and the pair of metrics $(\llvert \rho(B)^{-1}\rrvert _{1,\off}, \llvert \rho (B)^{-1}\rrvert _1)$ affect
that for the FF correspondent in~\eqref{eq::A-MLEcorr} as shown in
Theorem~\ref{thm::tildeAcorr}.
The performances in terms of edge recovery for $\Pi$ take a different order.
The sparse random graphs with $d = 90$ see better performances than
those with $d= 180$ for both the Gemini and the FF methods.
For graphs with the same sparsity, the one with the larger weight
performs better.
This is consistent with our theory in Section~14.1.\looseness=1

 \begin{table}
\caption{Metrics for comparing the ROC curves}\label{tab2}
\begin{tabular*}{\textwidth}{@{\extracolsep{\fill}}lcccc@{}}
\hline
&\multicolumn{1}{c}{$\bolds{d=90}$}& \multicolumn{1}{c}{$\bolds{d=180}$}& \multicolumn{1}{c}{$\bolds{d=90}$} & \multicolumn{1}{c@{}}{$\bolds{d=180}$} \\
\textbf{Metric} & $\bolds{w:[0.1,0.3]}$ & $\bolds{w:[0.1,0.3]}$ & $\bolds{w:[0.6,0.8]}$ & $\bolds{w:[0.6,0.8]}$ \\
\hline
$\rho_{B}^2$ & 0.053 & 0.06 & 0.094 & 0.12 \\
$\Vert{B}\Vert_F/\tr(B)$ & 0.128 & 0.13 & 0.155 & 0.16 \\
$\ell_1$-metrics & (55, 152) & (71, 166) & (99, 225) & (102, 216) \\
\hline
\end{tabular*}
\end{table}

Next, we choose two covariance matrices for both $A$ and $B$:
for $B$, we choose the two cases with different edge weights with
$d=180$; and for $A$, we set the parameter $\rho$ to $0.5$ or $0.7$.
The metrics for the two choices of $A$ are:
for $\rho=0.5$, we have $\rho_{A}^2 = 0.04$, $\llVert  A\rrVert _F/\tr(A) =
0.065$, and
$\ell_1$-metrics${} = (532, 1198)$. The corresponding numbers for $\rho
= 0.7$ are:
$0.07, 0.085$, and $(1095, 2262)$, respectively.

First, we note that the two cases of $B$ show the same trend
when $A$ is fixed.
In the right bottom two plots in Figure~\ref{fig:roc-AB},
for the graphs of $\Omega$, we find it easier to estimate when their
covariance matrices come with parameter $\rho= 0.7$, which results in
larger $\ell_1$ metrics, and hence larger weights on the inverse chain graph;
for the graphs of $\Pi$, we observe relatively larger performance
gains when $\rho= 0.5$
for $A$, with the most significant occurring when $w \in[0.1, 0.3]$
for $\Pi$, where
both $\rho(A)^{-1}$ and $\rho(B)^{-1}$ have smaller $\ell_1$ metrics
and the total
correlation $\rho_{B}^2 = 0.06$ is also small.
The least improvement we see occurs in case all three metrics are
large: $\rho= 0.7$, $w \in[0.6, 0.8]$, and $\rho_{B}^2 = 0.12$.
These findings are consistent with results in
Theorem~\ref{thm::tildeAcorr}
and~\ref{thm::tildeBcorr},
where we explicitly show the influence of the pairs of
$\ell_1$-metrics on the error bounds for the FF sample correlation estimates.

\subsection{Application to EEG data}
\label{subsec::eeg}

\begin{figure}

\includegraphics{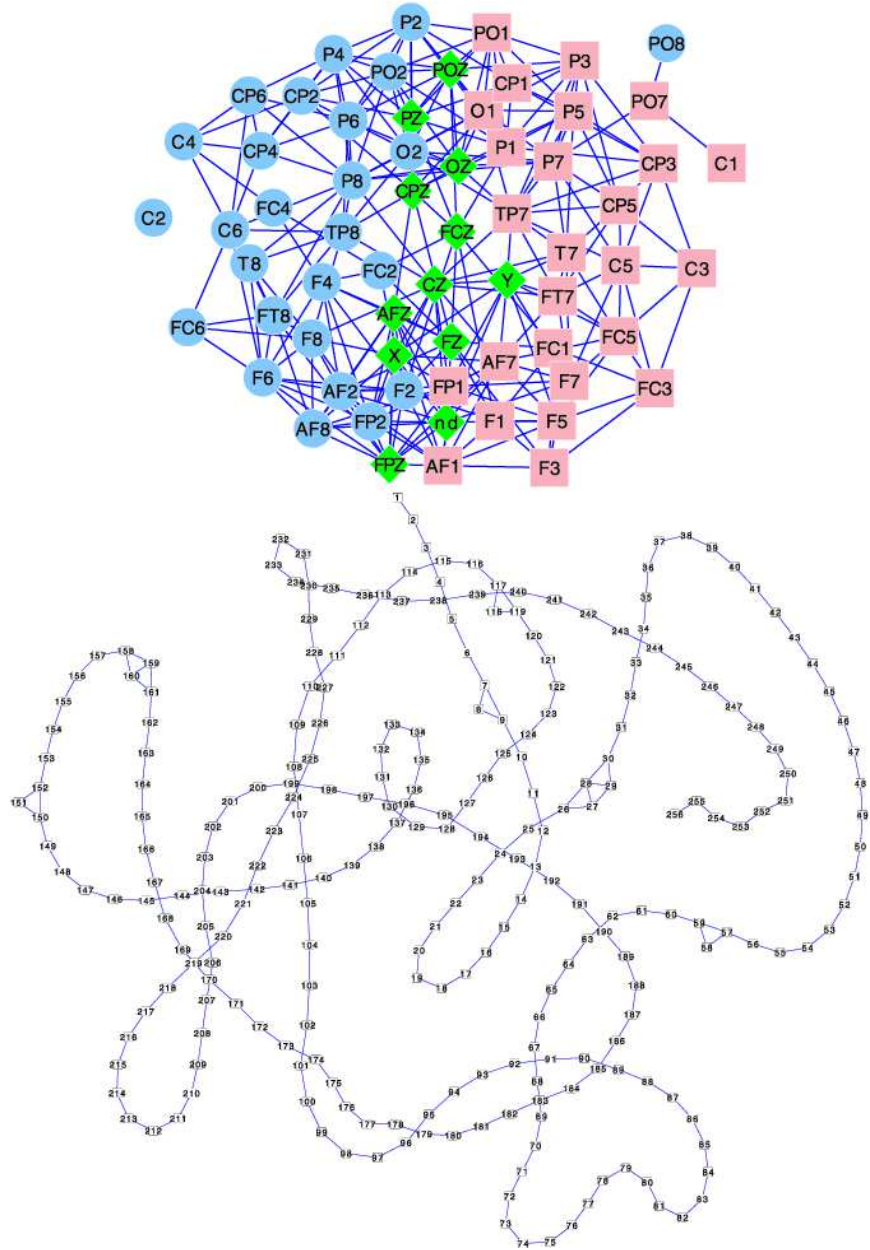}

\caption{\emph{Top}: Estimated graph of channels with penalty $
\lambda=0.40$.
Nodes are labeled with EEG electrode identifiers.
Circles, squares and diamonds represent electrodes placed
on the left, right, and middle of the head respectively. The graph
structure
indicates that nodes interact mostly with nodes that are physically
close to them.
\emph{Bottom}:
Estimated graph among measurements with penalty $\nu= 0.78$.
Nodes are labeled with epochs from 1 to 256.
The graph is primarily a long chain
showing sequential dependences among epochs with a few extra edges
between some neighbors.}
\label{fig:eeg-control-1}
\end{figure}

In this section, we present results of applying Gemini on real data. We used
the EEG (electroencephalography) data available from the UCI Machine learning
repository \cite{UCIeeg}, which was collected as part of the COGA
(Collaborative Studies on Genetics of Alcoholism) project \cite
{Zhang95}. The
data set we used contains measurements from 64 electrodes (channels)
placed on
two subjects' (one alcoholic and one control) scalps, which were
sampled at 256
Hz (3.9-msec epoch) for 1 second. The data consists of 10 runs under three
different stimulus paradigm. For each paradigm, we construct an $f
\times m $
matrix, $X$, for each subject's each run, where $f=64$ and $m=256$.
Each row in
$X$ represents a channel and each column represents a measurement
epoch. We
normalize each row vector such that its mean is 0 and variance is 1.
The 10
runs are treated as 10 replicates, and fed to Gemini to estimate both the
dependence structures of channels and measurements. We show the
resulting graphs
for control subject \texttt{c02c0000337} under one stimulus paradigm in Figure~\ref{fig:eeg-control-1}. The estimated graph among channels
largely reflects the spatial organization of the brain, and the
estimated graph
among measurement epochs suggests relatively short-order serial dependence.

\section{Conclusion}
\label{sec::conclude}
In this paper, we presented two methods for estimating graphs in a
matrix variate normal model. The baseline Gemini method is rather
simple and provides the same rates of convergence as the Noniterative
Penalized Flip-Flop method in the operator and the Frobenius norm.
In Gemini, a unique pair of optimal solutions
for the correlation matrices and their inverses are obtained via the
graphical Lasso algorithm. Under sparsity constraints and upon
multiplication by proper
weight matrices, the penalized estimators are strikingly
effective in approximating the row and column covariance matrices.
Under sparsity conditions as detailed in (A1) and (A2), the NiPFF
method shows some improvement over the baseline algorithm in estimating
$A_0^{-1}$, which is assumed to be the one with the larger dimension,
so long as $\rho(B_0)^{-1}$ satisfies a certain additional sparsity
condition, namely, its vector $\ell_1$ metrics are bounded in the
order of its dimensionality.
However, we show in both theoretical analysis and
simulation results that the performance gains for estimating
$B_0^{-1}$ using the NiPFF method at the third step are rather
limited; hence, we do not advocate iterating beyond the first three steps.
Although our primary interests are in estimating correlations and partial
correlations among and between both rows and columns when
$X$ follows a matrix variate normal distribution, our methods clearly
can be extended to the general cases when
the data matrix $X$ follows other type of matrix-variate
distributions.

\section*{Acknowledgements}
The author is grateful for the helpful discussions
with Xuming He, John Lafferty, Mark Rudelson, Kerby Shedden and
Stanislaw Szarek. The author thanks the Co-Editor Runze Li, an
Associate Editor and the anonymous referees for their valuable
comments and suggestions.



\begin{supplement}[id=suppA]
\stitle{Supplementary material for ``Gemini: Graph estimation with
matrix variate
normal instances''}
\slink[doi]{10.1214/13-AOS1187SUPP} 
\sdatatype{.pdf}
\sfilename{aos1187\_supp.pdf}
\sdescription{The technical proofs are given in the supplementary material~\cite
{Zhou13supp}.}
\end{supplement}

%

%

\printaddresses

\end{document}